\documentclass[10pt,a4paper,twocolumn]{article}

\usepackage[utf8]{inputenc}
\usepackage[T1]{fontenc}
\usepackage{newcent}  
\usepackage{amsmath,amssymb,amsfonts}
\usepackage{graphicx}
\usepackage[dvipsnames]{xcolor}
\usepackage[english]{babel}

\usepackage{geometry}
\usepackage[expansion=false]{microtype}
\usepackage{listings}
\usepackage{caption}
\usepackage{subcaption}

\DeclareCaptionFont{captionfont}{\fontsize{8pt}{10pt}\selectfont}
\usepackage{booktabs}
\usepackage{longtable}
\usepackage{multirow}
\usepackage{array}
\usepackage{tabularx}
\usepackage{float}
\usepackage{placeins}
\usepackage{stfloats}  
\usepackage{textcomp}
\usepackage{balance}
\usepackage{pdflscape}
\usepackage{appendix}

\usepackage[authoryear,round]{natbib}

\usepackage{tcolorbox}
\tcbuselibrary{skins,breakable}
\usepackage{enumitem}

\usepackage[hyphens,spaces,obeyspaces]{url}
\usepackage{hyperref}
\hypersetup{
    colorlinks=true,
    linkcolor=NavyBlue,
    citecolor=NavyBlue,
    urlcolor=NavyBlue,
    breaklinks=true
}

\makeatletter
\g@addto@macro{\UrlBreaks}{\UrlOrds}
\makeatother

\newcommand{\tablefontsize}{\fontsize{8pt}{10pt}\selectfont}

\definecolor{warmtitle}{HTML}{c0684d}
\definecolor{warmback}{HTML}{f4ddcf}
\definecolor{cooltitle}{HTML}{4da6c2}
\definecolor{coolback}{HTML}{d0e6f4}

\newtcolorbox{analogybox}[1][]{
    colback=warmback,
    colframe=warmtitle,
    coltitle=white,
    fonttitle=\bfseries,
    title={Computer Graphics Analogy},
    #1
}

\newtcolorbox{eli5box}[1][]{
    colback=coolback,
    colframe=cooltitle,
    coltitle=white,
    fonttitle=\bfseries,
    title={ELI5 (Explain Like I'm Five) for Engineers},
    #1
}

\newtcolorbox{keyinsight}[1][]{
    colback=coolback,
    colframe=cooltitle,
    coltitle=white,
    fonttitle=\bfseries,
    title={Key Insight},
    #1
}

\newtcolorbox{worldmodelbox}[1][]{
    colback=coolback,
    colframe=cooltitle,
    coltitle=white,
    fonttitle=\bfseries,
    title={World Model Connection},
    #1
}

\newtcolorbox{designerbox}[2][]{
    colback=warmback,
    colframe=warmtitle,
    coltitle=white,
    fonttitle=\bfseries,
    title={For Fashion Designers: #2},
    #1
}

\newtcolorbox{developerbox}[1][]{
    colback=coolback,
    colframe=cooltitle,
    coltitle=white,
    fonttitle=\bfseries,
    title={For ML/CS Practitioners},
    #1
}

\newtcolorbox{examplebox}[1][]{
    colback=warmback,
    colframe=warmtitle,
    coltitle=white,
    fonttitle=\bfseries,
    title={Concrete Example},
    #1
}

\newtcolorbox{verificationbox}[1][]{
    colback=coolback,
    colframe=cooltitle,
    coltitle=white,
    fonttitle=\bfseries,
    title={Verification Insight},
    #1
}

\newtcolorbox{techbox}[1][]{
    colback=warmback,
    colframe=warmtitle,
    coltitle=white,
    fonttitle=\bfseries,
    title={For Technical Practitioners},
    #1
}

\newtcolorbox{fashionconceptsbox}[1][]{
    colback=warmback,
    colframe=warmtitle,
    coltitle=white,
    fonttitle=\bfseries,
    title={Core Insight: Fashion as Constraint Satisfaction},
    #1
}

\newtcolorbox{programsynthesisbox}[1][]{
    colback=coolback,
    colframe=cooltitle,
    coltitle=white,
    fonttitle=\bfseries,
    title={Core Insight: From Pixels to Proofs},
    #1
}

\title{Textile IR: A Bidirectional Intermediate Representation for Physics-Aware Fashion CAD}

\author{Petteri Teikari$^{1,*}$, Neliana Fuenmayor$^{1}$ \\
\small $^{1}$Open Mode, London, United Kingdom \\
\small $^{*}$Corresponding author: petteri.teikari@gmail.com}

\date{\today}

\begin{document}

\maketitle

\begin{abstract}
We introduce Textile IR, a bidirectional intermediate representation that connects manufacturing-valid CAD, physics-based simulation, and lifecycle assessment for fashion design. Unlike existing siloed tools where pattern software guarantees sewable outputs but understands nothing about drape, and physics simulation predicts behaviour but cannot automatically fix patterns, Textile IR provides the semantic glue for integration through a seven-layer Verification Ladder---from cheap syntactic checks (pattern closure, seam compatibility) to expensive physics validation (drape simulation, stress analysis). The architecture enables bidirectional feedback: simulation failures suggest pattern modifications; material substitutions update sustainability estimates in real time; uncertainty propagates across the pipeline with explicit confidence bounds. We formalise fashion engineering as constraint satisfaction over three domains and demonstrate how Textile IR's scene-graph representation enables AI systems to manipulate garments as structured programs rather than pixel arrays. The framework addresses the compound uncertainty problem: when measurement errors in material testing, simulation approximations, and LCA database gaps combine, sustainability claims become unreliable without explicit uncertainty tracking. We propose six research priorities and discuss deployment considerations for fashion SMEs where integrated workflows reduce specialised engineering requirements. Key contribution: a formal representation that makes engineering constraints perceptible, manipulable, and immediately consequential---enabling designers to navigate sustainability, manufacturability, and aesthetic tradeoffs simultaneously rather than discovering conflicts after costly physical prototyping.
\end{abstract}

\textbf{Keywords:} intermediate representation, garment CAD, physics simulation, lifecycle assessment, constraint satisfaction, program synthesis, digital product passport, uncertainty quantification, fashion engineering

\begin{figure}[t]
\centering
\includegraphics[width=\columnwidth]{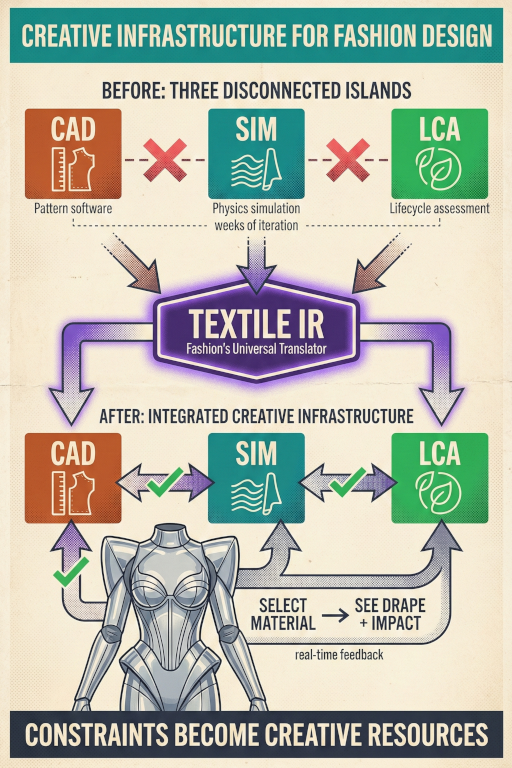}
\caption*{Graphical abstract: The Textile IR bridges three disconnected pillars---Manufacturing CAD, Physics Simulation, and Lifecycle Assessment---enabling bidirectional information flow for AI-assisted fashion engineering.}
\end{figure}

\section{Introduction}

\subsection{Islands of Automation}

Ask any technical designer about their last late-stage material failure. The story is depressingly familiar: a fabric specified for its sustainability credentials destroys the silhouette during sampling, requiring pattern redrafts, sustainability recalculations, and production delays. Generative AI was supposed to reshape fashion design~\citep{shi_genai_fashion_2025, rizzi_generative_2025, wu_ai_creativity_2025}---but it has not solved this problem because the problem is not generative capability. It is infrastructure.

Today's designer navigates disconnected islands of automation. Pattern software guarantees sewable outputs but understands nothing about drape; physics simulation predicts behaviour but cannot automatically fix patterns; lifecycle assessment estimates impact only after design decisions are locked. Each tool solves part of the problem, and none talks to the others. This is not a technical limitation waiting for better algorithms. It is a systematic human-computer interface (HCI) failure that wastes weeks and thousands of pounds per collection cycle.

Here's the frustrating part: all the necessary capabilities already exist. Manufacturing-valid CAD generates geometrically consistent patterns. Physics-based simulation predicts garment behaviour with increasing accuracy. Lifecycle assessment frameworks estimate environmental impact per standardised methodology. The gap is not capability---it is connection. We argue that integration is practical necessity, not a research luxury, and we propose a Unified Integration Framework that enables:

\begin{enumerate}
\item \textbf{A common language}: The ``Textile Intermediate Representation'' (Textile IR) enables bidirectional information flow---simulation failures suggest pattern modifications; material changes update sustainability estimates.
\item \textbf{Uncertainty as transparency}: Connected systems track how measurement errors compound, providing designers honest confidence bounds rather than false precision \citep{bhatt_uncertainty_2021}.
\item \textbf{Invisible engineering}: Human-in-the-loop interfaces present constraints as natural creative boundaries rather than technical obstacles. Abstraction levels vary with expertise: novices benefit from abstracted feedback; experts require direct constraint manipulation. Empirical studies should validate these intuitions.
\end{enumerate}

This framework particularly benefits fashion SMEs, where integrated workflows reduce the need for specialised engineering roles \citep{langley_orchestrating_2023}.

The Green Claims Directive (EU 2024/825) adds regulatory urgency. Claims with hidden uncertainty may qualify as greenwashing regardless of intent.

\subsection{The Design-Engineering Divide}

The integration gap compounds an older problem. Designers manage aesthetics. Engineers manage manufacturing. The handoff between them delays production and multiplies revision cycles \citep{watkins_functional_2015, tyler_supply_2006, parker-strak_challenges_2023}. This separation is organisational, not technical.

The stakes are high. Fashion accounts for 8--10\% of global greenhouse gas (GHG) emissions \citep{niinimaki_environmental_2020, filho_overview_2022}. The EU's Ecodesign for Sustainable Products Regulation (ESPR), in force since July 2024, mandates Digital Product Passports (DPPs)~\citep{eu_parliament_dpp_2024}. The secondhand market's projected growth to \$317 billion by 2027 \citep{mckinsey_state_2025} reframes DPPs from compliance burden to business enabler \citep{langley_orchestrating_2023}. Compliance and competitive advantage are converging~\citep{reich_beyond_2025}.

\subsection{Creativity and Constraints}
\label{sec:creativity}

The obvious objection: constraints limit creativity. We argue the opposite.

Regulatory and market pressures could be addressed through compliance-oriented solutions~\citep{reich_beyond_2025, tamm_integrating_2026}. But we propose a deeper reframing. Design practice reveals that constraints do not merely restrict creative outcomes---they shape them. Recent work on AI-assisted fashion creativity~\citep{wu_ai_creativity_2025, shi_genai_fashion_2025} demonstrates that generative tools reshape the designer's iterative dialogue: proposing variations, receiving feedback, refining intent. Current disconnected tools impoverish this dialogue. Physics behaviour and environmental impact remain invisible until physical samples reveal problems---weeks later.

The Textile IR changes this conversation. Select a material: see predicted drape behaviour immediately. Modify a pattern: sustainability estimates update instantly. Designers explore ``what if'' scenarios without physical commitment. They develop intuitions about constraint interactions that sequential workflows cannot support.

The reframing is fundamental. Engineering constraints, when made visible, manipulable, and immediately consequential, become creative resources---not limitations~\citep{langley_orchestrating_2023}. This inversion underpins everything that follows.

\subsection{The Promise and Limitation of Current AI}

Generative AI has undeniably transformed early-stage design exploration---text-to-image systems enable mood board ideation at unprecedented speed~\citep{shi_genai_fashion_2025}, while neural style transfer~\citep{zhu_unpaired_2017} and virtual try-on~\citep{islam_deep_2024} accelerate visualisation. But there is a catch. A comprehensive review of deep learning for 3D garment generation~\citep{sun_deep_2025} puts it bluntly: ``integration of simulation, pattern semantics, and sustainability remains unexplored.''

The outputs lack structure---generative AI produces raster images, pixels without pattern data, manufacturing specifications, or sustainability metrics. A designer cannot take a Midjourney image and send it to a factory. DressCode \citep{he_dresscode_2024} demonstrates impressive text-to-garment capabilities, yet larger training sets in their context produced outputs that violate manufacturing constraints more frequently---the ``data-quality paradox.'' Recent work addresses ``structural hallucinations''~\citep{li_ragdiffusion_2025}---outputs appearing plausible but containing geometric impossibilities. Neural networks generating patterns learn co-occurrence, not constraint satisfaction \citep{chollet_measure_2019}.

\subsection{The Problem Evolution: 2020--2025}

GarmentCode \citep{korosteleva_garmentcode_2023} introduced domain-specific languages (DSLs) guaranteeing geometric validity by construction. The 2025 convergence toward physics-constrained generation is evidenced by systems coupling neural generation with simulation validation: DSO aligns 3D generators with physics feedback~\citep{yang_dso_2025}; D-Garment conditions diffusion models on dynamic physics~\citep{zhang_dgarment_2025}; Dress-1-to-3 produces simulation-ready outputs from single images~\citep{li_dress-1-to-3_2025}.

Yet no system fully closes the loop from text prompt or visual sketch through physics-valid simulation to verifiable sustainability claim~\citep{tailor_2025}. This is our central thesis: the Textile IR connecting CAD, simulation, and LCA enables reliable fashion engineering that isolated tools cannot achieve. The parallel to IC design is instructive: semiconductor manufacturers adopted shift-left methodologies to catch design flaws before expensive fabrication \citep{wu_shiftleft_2025}. Google DeepMind's FunSearch \citep{romera-paredes_mathematical_2024} demonstrates the paradigm: LLMs propose, formal systems verify.

\subsection{Why Integration Matters: The Cascade of Disconnection}

\begin{sloppypar}
Here is what happens. A designer selects organic cotton for sustainability. Develops patterns in CAD. Sends for fit validation. Physics simulation reveals the cotton's drape creates an unflattering silhouette. The choices: accept the compromised silhouette, redesign (weeks of work), or switch materials (invalidating the sustainability rationale). None is good.
\end{sloppypar}

Why? CAD, simulation, and LCA operate on incompatible representations. No shared semantic layer. Each iteration requires weeks and physical samples \citep{gutin_interdiction_2015, parker-strak_challenges_2023}. The disconnection is structural, not accidental.

\subsection{Scope and Limitations}

We are explicit about scope. This article presents an architectural framework---the Textile IR specification and Verification Ladder---synthesising capabilities demonstrated separately in published systems. Computational feasibility of individual components (GarmentCode, MPMAvatar, PEF methodology) has been validated in their respective publications; their integration awaits implementation. Whether designers prefer constraint visibility over abstraction requires empirical validation \citep{dove_design_2017, amershi_guidelines_2019}. Organisational barriers receive limited attention \citep{bertola_fashion_2018}. These gaps define the research agenda in Section~7.

\subsection{Research Questions}

This article addresses: (1) How can AI-generated patterns guarantee manufacturing validity? (2) How can physics simulation validate fit across body diversity? (3) How can LCA integrate into real-time material selection? (4) What research agenda enables integration?

\subsection{Contribution and Structure}

Three contributions. First, we formalise the integration requirements for manufacturing CAD, physics simulation, and LCA---identifying data representation gaps and proposing interchange semantics. Second, we introduce the Textile IR concept---an intermediate representation enabling bidirectional information flow analogous to compiler IRs. Third, we reframe fashion engineering as a program synthesis problem. The typical gradient-based optimisation cannot navigate topology-dependent design spaces---you cannot smoothly ``slide'' from a two-piece to a three-piece pattern. Our central argument: disconnected tools create compound uncertainty that renders sustainability claims statistically unreliable. A designer sees ``Carbon: 12.3 kg CO$_2$e'' when the honest answer is ``Carbon: 12.3 $\pm$ 3.2 kg CO$_2$e (95\% CI)''---a range so wide that material comparisons become meaningless.

\section{The Convergence Problem: Why Each Pillar Alone Fails}

Here is a frustrating scenario that plays out every season. GarmentCode \citep{korosteleva_garmentcode_2023} guarantees your pattern is sewable---geometrically valid by construction. MPMAvatar \citep{lee_mpmavatar_2025} tells you how the garment will drape. FAHP-TOPSIS \citep{chow_design_2005} helps you weigh sustainability trade-offs. Three excellent tools. Three separate workflows. Three weeks before you discover the pattern is sewable, the drape is wrong, and the sustainability numbers need recalculating.

Why does this keep happening? Because each pillar creates dependencies the others cannot see. A manufacturing-valid pattern without physics validation produces garments that are technically correct but experientially wrong---they fit badly, they drape like cardboard, they fail in ways that only emerge during sampling. Physics simulation without parametric CAD cannot generalise to new designs. LCA without CAD and simulation produces numbers disconnected from design decisions. The tools do not talk to each other, and we pay for that silence in wasted time and materials.

\subsection{Industry Momentum Beyond Academic Publication}

Commercial tools move faster than peer-reviewed literature. CLO3D, Browzwear, The Fabricant---integration concepts are already deployed. Practitioner interviews confirm selective adoption: designers use CLO3D and Browzwear primarily for early visualisation and digital prototyping, but full integration remains limited by organisational, technical, and cultural barriers~\citep{selkee_ai-driven_2025}. Partial integration is not integration. CLO3D incorporates sustainability estimation. Browzwear offers PLM integration. Neither provides the bidirectional flow we argue is necessary.

The Textile IR contribution is not replacing these capabilities. It is the semantic layer enabling cross-tool interoperability. When we cite GarmentCode or MPMAvatar, we cite peer-reviewed foundations. Commercial implementations may exceed documented capability. We do not know. Academic publication lags practice.

Architecture provides instructive precedent: building information modelling (BIM)-integrated lifecycle assessment tools now deliver real-time sustainability feedback during design~\citep{ma_bim_lca_2025}, demonstrating that fashion's proposed CAD-simulation-LCA integration is technically feasible rather than speculative.

\begin{figure*}[htbp!]
\centering
\includegraphics[width=\textwidth]{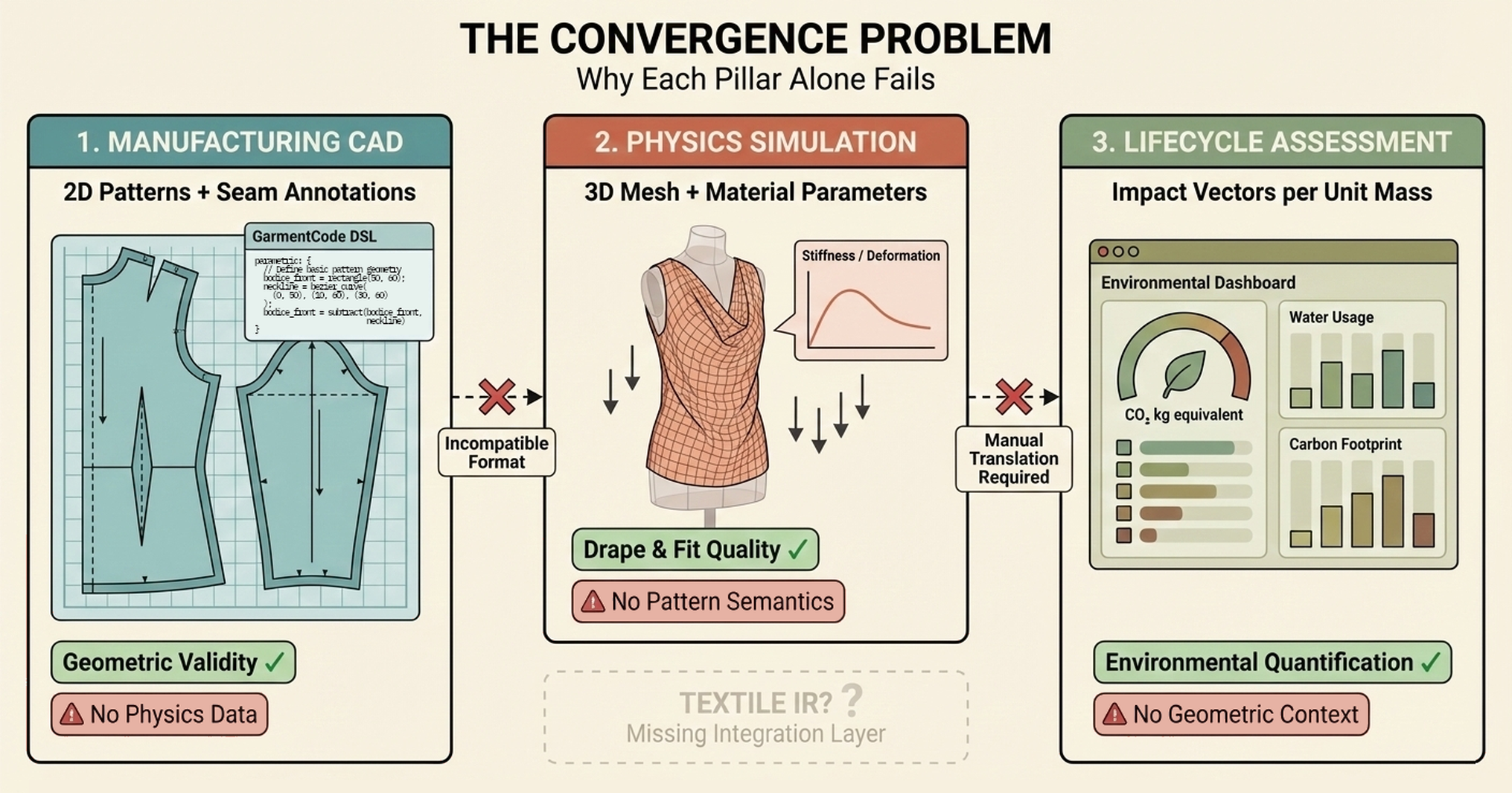}
\caption{The three pillars of AI-assisted fashion engineering operate as disconnected islands despite individual maturity. CAD guarantees geometric validity, physics simulation predicts drape, and LCA quantifies environmental impact. Each pillar's strength creates a blind spot: CAD produces patterns without physics data, simulation loses pattern semantics, LCA calculates impact without geometric context. No shared semantic layer enables bidirectional flow. }
\label{fig:architecture}
\end{figure*}

Table~\ref{tab:pillars} summarises each pillar's capabilities, while Figure~\ref{fig:architecture} visualises data flow relationships. Table~\ref{tab:commercial} compares commercial tool capabilities across the three pillars.

\begin{table*}[t]
\centering
\caption{Three pillars of AI-assisted fashion engineering: capabilities, limitations, and interdependencies.}
\label{tab:pillars}
\tablefontsize
\begin{tabularx}{\textwidth}{@{}lXXX@{}}
\toprule
\textbf{Aspect} & \textbf{Manufacturing CAD} & \textbf{Physics Simulation} & \textbf{Lifecycle Assessment} \\
\midrule
Representation & 2D patterns + seam annotations & 3D mesh + material parameters & Impact vectors per unit mass \\
Optimizes for & Geometric validity, marker efficiency & Experiential quality (drape, fit) & Environmental impact minimization \\
Key system & GarmentCode DSL \citep{korosteleva_garmentcode_2023} & MPMAvatar \citep{lee_mpmavatar_2025} & PEF / PEFCR \\
Limitation & No physics---cannot predict drape & No pattern semantics---loses intent & No geometry---disconnected from design \\
Requires from others & Physics validation for fit & Pattern semantics for modification & Geometric context for per-product impact \\
Success rate & Geometric validity by construction; $\sim$72\% simulation convergence \citep{korosteleva_garmentcodedata_2024} & Zero-shot generalisation to unseen geometries & Per-unit accurate within methodology \\
\bottomrule
\end{tabularx}
\end{table*}

\begin{table*}[t]
\centering
\caption{Commercial fashion design tools: capabilities across the three pillars. SMPL = Skinned Multi-Person Linear model.}
\label{tab:commercial}
\tablefontsize
\begin{tabularx}{\textwidth}{@{}lccccX@{}}
\toprule
\textbf{Tool} & \textbf{Physics} & \textbf{LCA} & \textbf{Body Diversity} & \textbf{CAD Export} & \textbf{Integration Gap} \\
\midrule
CLO3D & High & Partial & SMPL-based & DXF & No bidirectional LCA flow \\
Browzwear & High & PLM link & Avatar library & DXF/AAMA & Manual sustainability input \\
Optitex & Medium & None & Size grading & DXF & Physics accuracy unvalidated \\
Style3D & High & None & Custom avatars & Proprietary & Closed ecosystem \\
The Fabricant & Rendering & None & Limited & None & Visualisation only \\
GarmentCode (research) & Simulation-ready & None & None & Parametric & No LCA, no material physics \\
\bottomrule
\end{tabularx}
\end{table*}

\subsection{Manufacturing-Valid Pattern Generation}

Fashion's digital transformation proceeded in waves. First: 2D CAD systems digitising pattern drafting \citep{liu_interactive_2018, chaudhary_maximizing_2020}, with recent evaluations showing body-shape customisation requires substantial patternmaking expertise~\citep{guo_evaluation_2023}. Then: 3D visualisation promising virtual fit. Now: generative AI enabling text-to-image exploration. Yet each wave failed to close the production-ready gap. A 3D visualisation showing perfect fit may derive from patterns that cannot be assembled---the core problem being that neural pattern generators learn probability distributions, not manufacturing constraints.

GarmentCode \citep{korosteleva_garmentcode_2023} takes a fundamentally different approach: patterns expressed as programs with constraints satisfied by construction, achieving geometric validity by definition. The GarmentCodeData dataset \citep{korosteleva_garmentcodedata_2024} demonstrates that 72\% of these geometrically-valid patterns also achieve physics simulation convergence across 115,000 diverse garments; the remaining 28\% fail simulation due to self-intersection or numerical instability---not geometric invalidity. Importantly, simulation convergence does not guarantee drape accuracy, which requires validated material parameters. Design2GarmentCode \citep{zhou_design2garmentcode_2025} extends to multimodal input---neural translation to symbolic representation.

AIpparel~\citep{nakayama_aipparel_2025} scales this approach with a 7B parameter foundation model. ChatGarment~\citep{bian_chatgarment_2025} enables designers to refine patterns through natural language dialogue. Related work includes Neural Sewing Machines \citep{chen_structurepreserving_2022} and NeuralTailor \citep{korosteleva_neuraltailor_2022}; industry evaluation \citep{chen_automating_2025} reports 42\% usability with focused training. Yet a pattern can be geometrically valid while producing poor fit---manufacturing-valid CAD therefore requires physics-based simulation.

\subsection{Physics-Based Simulation}

Virtual try-on promises to reduce physical sampling \citep{song_imagebased_2023, islam_deep_2024}. The field has moved from mass-spring models \citep{baraff_large_1998} through finite element methods \citep{kim_finite_2020, stuyck_cloth_2018, chen_learning_2025} to neural radiance fields (NeRFs) and Gaussian splatting \citep{balloni_neural_2025, jiang_avatarvton:_2025}. Impressive quality. Missing structure.

\begin{figure*}[htbp!]
\centering
\includegraphics[width=\textwidth]{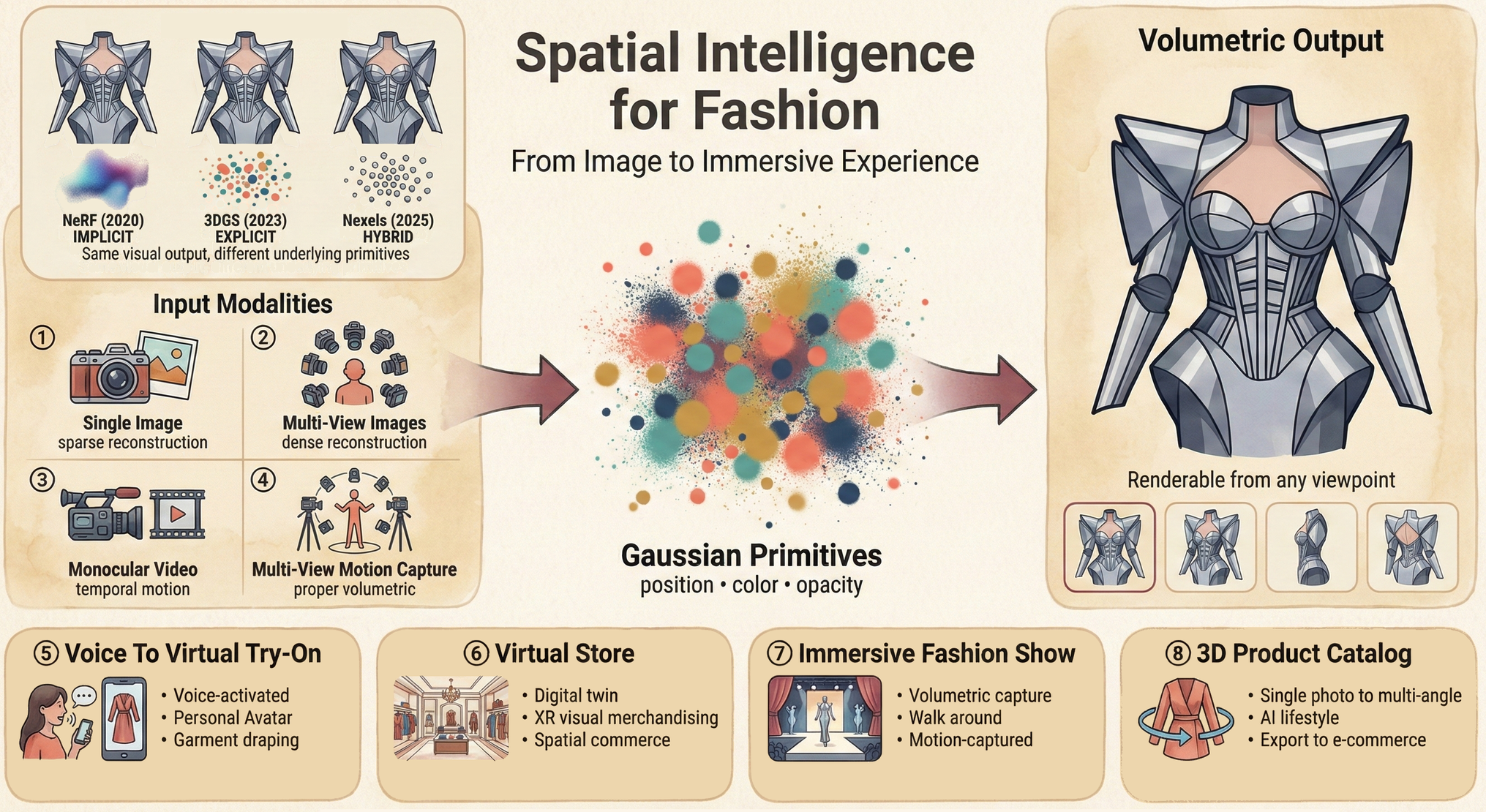}
\caption{\textbf{Spatial intelligence converts fashion imagery into immersive experiences.} Neural radiance fields~\citep{mildenhall_nerf_2020} pioneered neural scene representation (2020), evolving through 3D Gaussian Splatting~\citep{kerbl_3d_2023} (2023) to hybrid neural-geometric representations. Modern world models accept diverse inputs from single photographs to multi-view capture. Fashion applications include virtual try-on, digital twin stores, and immersive fashion shows. }
\label{fig:spatial-intelligence}
\end{figure*}

3D Gaussian Splatting (3DGS) \citep{kerbl_3d_2023, balloni_neural_2025} enables real-time photorealistic rendering (Figure~\ref{fig:spatial-intelligence}). Recent advances address self-collision and material anisotropy~\citep{ban_sagsgnn_2025}. A critical gap remains: these systems excel at appearance but do not guarantee physics. The research question has shifted from ``can we render realistically?'' to ``can we connect rendering to physics?''

The deeper problem: generalisation. Learned approaches memorise deformation patterns. They cannot extrapolate beyond training. MPMAvatar \citep{lee_mpmavatar_2025} takes a different path---physics-based simulation using Material Point Method, achieving zero-shot generalisation to unseen geometries. The key: similar material properties. Complementary advances include differentiable avatar methods \citep{chen_learning_2025}.

\textbf{Bridging Simulation to Textile Science.} Physics simulation parameters must connect to industry-standard fabric characterisation. The Kawabata Evaluation System (KES), widely adopted in textile engineering, defines measurable properties---bending rigidity (B), shear hysteresis (2HG), tensile linearity (LT)---that predict drape behaviour. Industry standards including ASTM D1388 (fabric stiffness) and AATCC Test Method 66 (wrinkle recovery) provide standardised testing protocols. The integration gap: simulation engines specify Young's modulus and Poisson's ratio, while material databases report Kawabata parameters. Establishing validated mappings between simulation parameters and textile metrology would enable designers to select fabrics by experiential properties (``flowing drape,'' ``structured hold'') while simulations operate on physical constants. Recent work demonstrates this integration is achievable: \citet{dominguez-elvira_practical_2024} show that fabric mechanical properties (stretch modulus, bending rigidity) can be estimated from manufacturer metadata alone---fabric family, density, composition, and thickness---achieving strong correlation with Cusick drape ground truth across 2,575 fabrics without expensive lab equipment. This ``metadata-to-mechanics'' bridge enables DPP data to directly inform simulation parameters.

Research on body-fit pattern generation \citep{oh_generation_2025} establishes that practitioners require tight fit tolerances. SMPL-based parametric body models \citep{loper_smpl_2015} enable systematic validation across body diversity.

\subsection{Lifecycle Assessment}

Fashion brands face pressure to substantiate sustainability claims \citep{navarrete_flue_2021}, yet practitioners remain ``lost in a sea of specialized knowledge'' when navigating sustainability indices~\citep{palomo-lovinski_missed_2024}. Comprehensive LCA remains expensive, slow, disconnected from design decisions. The question: what if LCA informed material selection at design time? We call this ``Shift-Left'' for physical production.

The EU PEF methodology (PEFCR version 3.1, May 2025) provides standardised framework with 16 environmental impact categories \citep{dhiwar_life_2025}. LCA requires defining functional units to enable meaningful comparisons. Durability directly affects denominators---technical quality represents lifespan potential~\citep{aakko_managing_2024}. A garment lasting 300 wears halves per-wear impact compared to one lasting 150 wears. Carbon footprint values here assume cradle-to-gate boundaries per PEFCR methodology.

\begin{figure*}[htbp!]
\centering
\includegraphics[width=\textwidth]{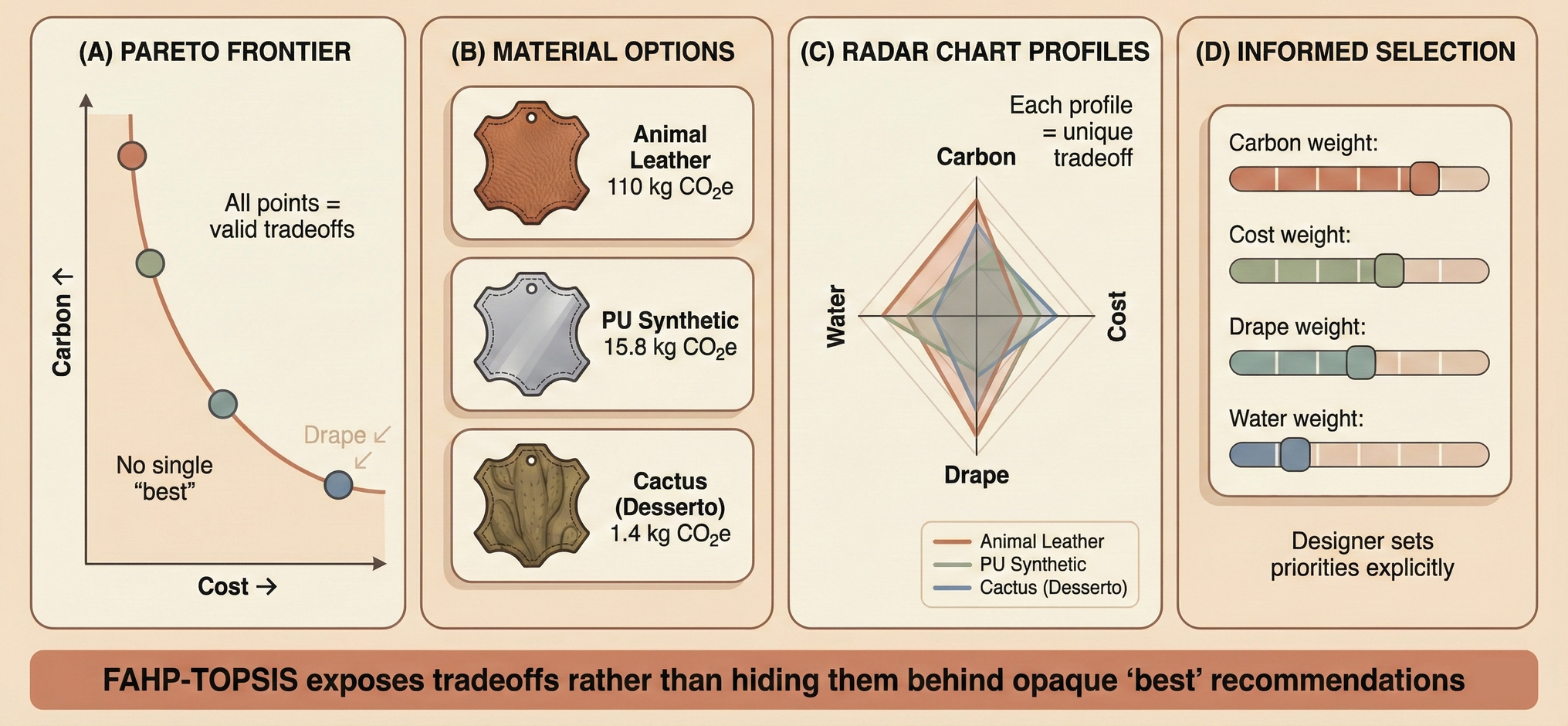}
\caption{\textbf{Multi-criteria decision making exposes tradeoffs rather than hiding them.} (A) Pareto frontier showing leather material options. (B) Material options with LCA data: animal leather (110~kg~CO$_2$e/m$^2$), PU synthetic (15.8~kg~CO$_2$e/m$^2$), cactus leather (1.4~kg~CO$_2$e/m$^2$). (C) Radar profiles showing unique performance across criteria. (D) FAHP-TOPSIS weighting interface. }
\label{fig:mcdm-tradeoff}
\end{figure*}

\textbf{DPP as Design Input, Not Output.} Current framings position Digital Product Passports as compliance outputs---documentation created after design completion. We argue the opposite: DPP requirements should function as design inputs. Emerging architectural frameworks reframe DPPs from compliance burdens to value-creating assets through AI-driven data quality and interoperability~\citep{tamm_integrating_2026}; verification hierarchies can decompose from automated checks through to human expert review. Multi-criteria decision-making for sustainability trade-offs has precedent in supply chain management~\citep{govindan_multi_2015}. FAHP-TOPSIS~\citep{chow_design_2005} enables ranking options across multiple criteria with uncertain weights (Figure~\ref{fig:mcdm-tradeoff}).

A critical challenge: efficiency gains can paradoxically increase environmental impact through rebound effects \citep{santos_circularity_2025}. Secondhand consumption positively correlates with new purchases \citep{nature_secondhand_2025}. A sustainability co-pilot must incorporate sufficiency guidance, not merely optimise efficiency metrics \citep{bocken_sufficiency_2022}.

The DPP regulatory timeline creates urgency~\citep{eu_parliament_dpp_2024}: Phase 1 (2027) requires minimal DPP with LCA data; Phase 2 (2030) mandates enhanced traceability; Phase 3 (2033) demands full circular DPP. Brands have approximately 18 months from delegated act publication to implement Phase 1 compliance.

AI-assisted design systems generate reasoning chains that serve dual purposes: designer guidance and Digital Product Passport provenance~\citep{reich_beyond_2025, tamm_integrating_2026}. This reframes explainability from usability feature to compliance infrastructure \citep{bhatt_uncertainty_2021, marx_uncertainty_2023}.

LCA databases characterise materials by impact per unit mass, but garment impact depends on pattern geometry. Material selection affects both environmental impact and physical behaviour. Integration would enable design-aware LCA: environmental impact accounting for pattern geometry and material behaviour.

AI enables circular economy applications~\citep{ramirez-escamilla_advancing_2024}: sustainability assessment using deep learning~\citep{nisa_systematic_2025}, durability policy analysis~\citep{richardy_wicked_2025}, and waste management optimisation. System dynamics modelling demonstrates that prioritising consumption reduction measures proves more effective than solely increasing sorting capacity. Computational infrastructure carries environmental costs; efficiency gains emerge from model right-sizing, token-efficient orchestration, and semantic caching.

\section{Integration Barriers: Why Connection Is Hard}
\label{sec:integration}

Integration failed for reasons worth understanding. Four barriers block connection: incompatible data formats, conflicting objectives, computational cost, and absent shared ontology.

\subsection{Data Representation Incompatibilities}

\begin{figure*}[htbp!]
\centering
\includegraphics[width=\textwidth]{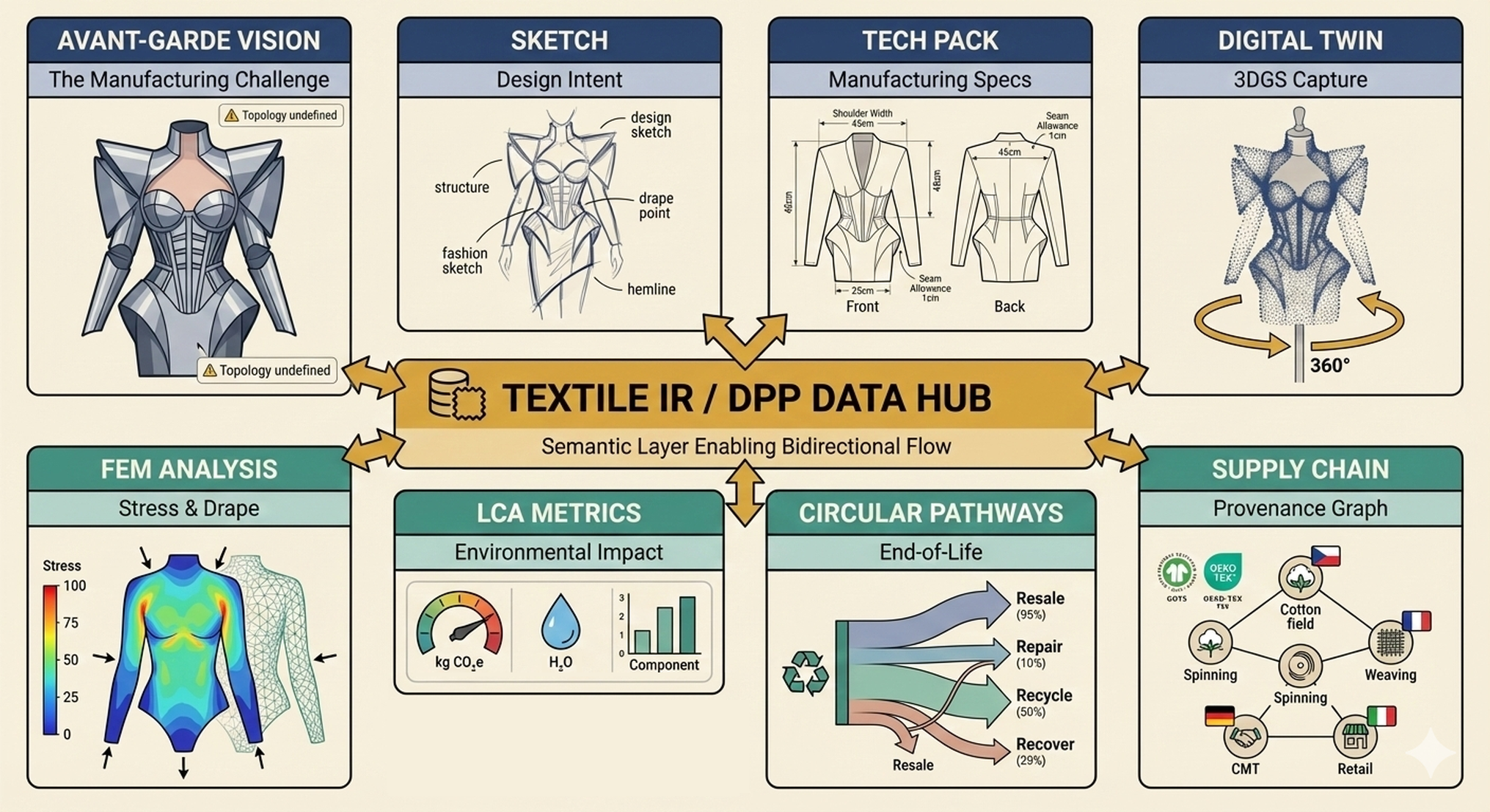}
\caption{Eight data representations spanning the fashion product lifecycle, connected through a central Textile IR / DPP Data Hub. Design representations (Vision, Sketch, Tech Pack, Digital Twin) prioritise creative intent. Simulation representations (FEM) prioritise physical accuracy. Lifecycle representations (LCA, Circular, Supply Chain) prioritise environmental metrics. Only the central hub can bridge these modalities. }
\label{fig:representations}
\end{figure*}

Fashion product development involves fundamentally different representations. Translation is lossy. Often manual. Sometimes impossible (Table~\ref{tab:incompatibilities}, Figure~\ref{fig:representations}).

\begin{table*}[t]
\centering
\caption{Data representation incompatibilities across pillars.}
\label{tab:incompatibilities}
\tablefontsize
\begin{tabularx}{\textwidth}{@{}lXXXX@{}}
\toprule
\textbf{System} & \textbf{Format} & \textbf{Contains} & \textbf{Lacks} & \textbf{Translation Path} \\
\midrule
Manufacturing CAD & GarmentCode DSL, DXF-AAMA & Geometry, seams, grainlines & Material physics, LCA data & Manual mesh conversion \\
Physics Simulation & 3D mesh + material params & Physics, deformation & Pattern semantics, LCA IDs & Metadata-to-mechanics~\citep{dominguez-elvira_practical_2024} \\
Lifecycle Assessment & Database entries + vectors & Environmental indicators & Geometry, physical behaviour & Manual specification \\
\bottomrule
\end{tabularx}
\end{table*}

\subsection{Conflicting Optimisation Objectives}

The pillars optimise for different things. Manufacturing CAD: geometric validity, production efficiency. Physics simulation: experiential outcomes. LCA: environmental impact~\citep{niinimaki_environmental_2020}. These conflict. A flowing silhouette requires drapeable fabric. Silk drapes beautifully but has high impact. Hemp has low impact but drapes like cardboard. No single pillar resolves multi-objective optimisation---requiring formal multi-criteria frameworks~\citep{govindan_multi_2015}.

\subsection{Computational Bottlenecks}

Physics simulation remains computationally demanding. MPMAvatar~\citep{lee_mpmavatar_2025}: $\sim$1.1 seconds per frame at standard resolution on research hardware. Interactive design workflows require sub-5-second feedback across multiple variants. The gap is not small.

Commercial tools use position-based dynamics (XPBD) for speed. Research systems favour finite element methods (FEM) for accuracy~\citep{stuyck_cloth_2018, kim_finite_2020}. Hierarchical architectures combining both await fashion-specific validation.

\subsection{Absence of Shared Ontology}

The three domains lack shared vocabulary. A ``seam'' in CAD: a geometric relationship. In simulation: a mechanical constraint. In LCA: a material consumption factor. Same word, different meanings. But DPP metadata provides a natural anchor. Fabric family, composition, density---the same attributes appear in manufacturer databases, simulation parameter estimation~\citep{dominguez-elvira_practical_2024}, and LCA inventories. Semantic web technologies could formalise these into comprehensive ontologies~\citep{aime_vetivoc_2016, pourjafarian_odp-based_2025}.

\subsection{Mathematical Hardness}

\begin{figure}[htbp!]
\centering
\includegraphics[width=\columnwidth]{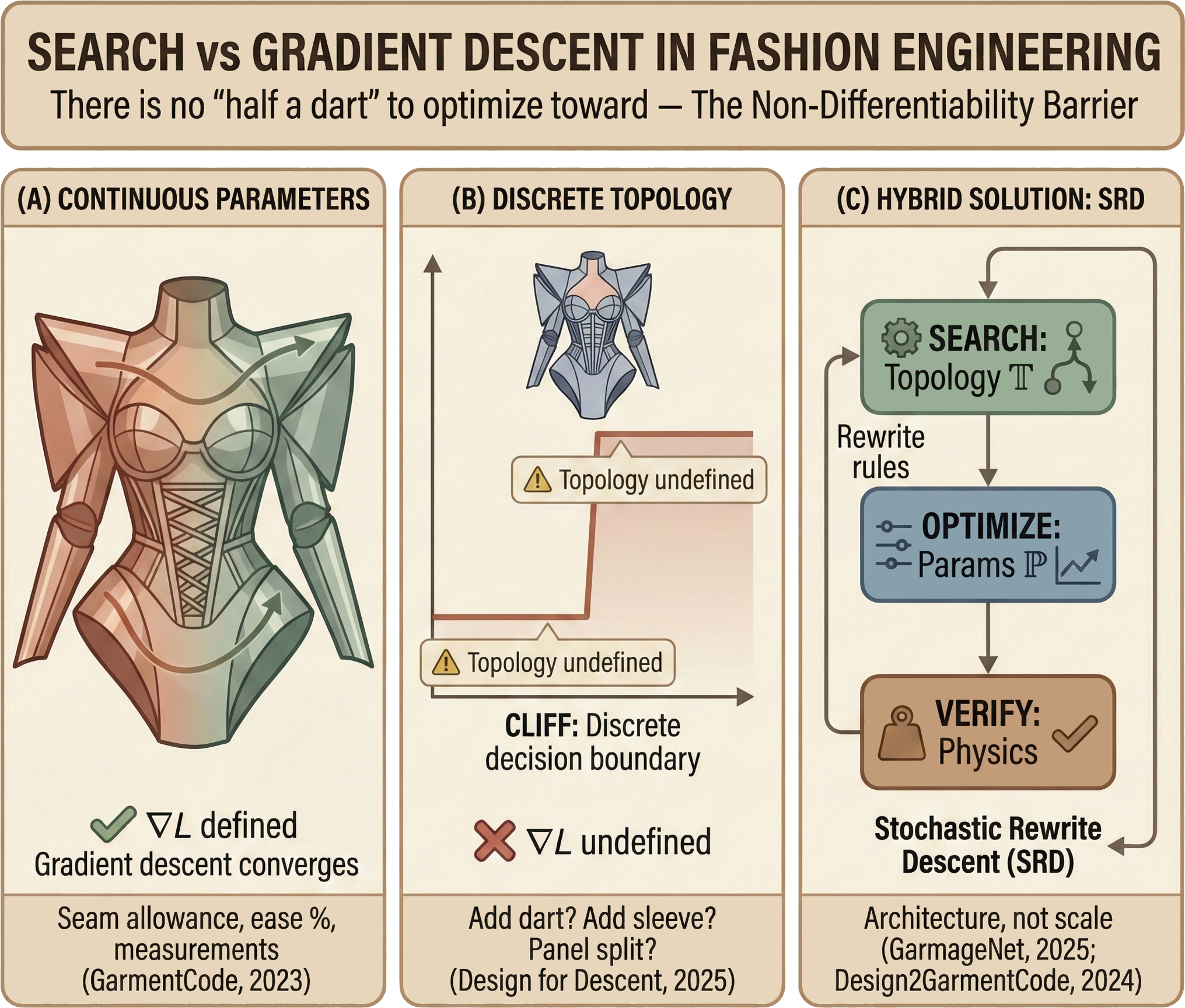}
\caption{\textbf{The non-differentiability barrier.} (A) Continuous parameters form smooth optimisation landscapes where gradient descent converges. (B) Discrete topology decisions create discontinuities---there is no ``half a dart.'' (C) Hybrid solution: search over topology, optimisation over parameters, then physics verification~\citep{korosteleva_garmentcode_2023, kodnongbua_design_2025}. }
\label{fig:search-gradient}
\end{figure}

Mathematical hardness explains integration's absence. Differentiable physics engines are maturing \citep{murthy_gradsim_2020}, and cloth simulation with complex self-collisions remains harder than rigid body cases. But the deeper problem: ``differentiating through the sewing pattern'' is impossible. When a designer adds a dart, the mesh topology changes discretely. Gradient descent cannot cross topological boundaries. You cannot smoothly ``slide'' from a two-piece to a three-piece pattern. At some point you must jump (Figure~\ref{fig:search-gradient}). Topology changes create discontinuities in the design-space manifold, violating the Lipschitz continuity required for gradient-based optimisation.

This decomposition---discrete search over structure, continuous optimisation over measurements---explains why scaling alone cannot close the gap.

\subsection{Search, Not Gradient Descent}
\label{sec:search-paradigm}

Here is the reframing. AI-assisted fashion engineering is a \emph{search problem}, not an optimisation problem. This distinction has profound implications for computational architecture. The distinction is conceptual, not merely technical. Image generation learns co-occurrence patterns. Training on diverse street fashion teaches that hoodies co-occur with kangaroo pockets. But co-occurrence does not encode the topological constraint that pocket openings must align with hand position when worn. A sleeve armhole requires convex curvature. The bodice armscye requires matching concave curvature. Neural networks trained on 2D pattern images have no explicit representation of ``seam edge.'' They learn probability distributions. Not constraint satisfaction \citep{chollet_measure_2019, faruqi_shaping_2024}.

Program synthesis inverts verification order. Instead of rendering first and checking validity afterward, specifications are verified against constraints \emph{before} any pixels are drawn. GarmentCode achieves 100\% simulation \emph{convergence}---not accuracy, which depends on material parameter correctness---because geometrically invalid patterns are rejected at parse time \citep{korosteleva_garmentcode_2023, zhou_design2garmentcode_2025}. They never reach the physics engine. This is not optimisation. It is a mathematical guarantee.

FunSearch \citep{romera-paredes_mathematical_2024} and AlphaTensor \citep{fawzi_discovering_2022} demonstrate the paradigm: LLM proposal generation combined with formal verification. The fashion analog: an LLM proposes pattern modifications, the GarmentCode DSL verifies geometric validity, physics simulation evaluates drape, and search algorithms guide iteration toward designs satisfying all constraints.

A critical consideration shapes architecture: physics simulation is expensive ($\sim$1.1s per frame), while formal verification of geometric constraints is cheap (milliseconds). The asymmetry matters. FunSearch-style architectures generate thousands of candidates; fashion search must filter most through fast geometric validation before expensive physics. Experience with LLM-verifier pipelines suggests orders of magnitude more proposals than accepted solutions~\citep{romera-paredes_mathematical_2024}. The implication: hierarchical verification---syntactic validity (instant), geometric validity (milliseconds), physics simulation (seconds), human approval (minutes). Cheap checks gate expensive operations, making design space exploration tractable.

\subsection{Textile IR: Fashion's Universal Translator}
\label{sec:textile-ir-spec}

If CAD, physics simulation, and LCA are the ``Islands of Automation,'' the Textile Intermediate Representation (Textile IR) is the high-speed ferry connecting them. Borrowing from compiler theory, the IR is not just a file format; it is a semantic graph that understands a ``seam'' is not just a line---it is a mechanical contract between two pieces of cloth.

Design space exploration is computationally intractable without filtering. A single physics simulation takes seconds to minutes. Exhaustive search across material, pattern, and sizing variations is impossible. The IR formalises the semantic layer while enabling hierarchical verification that makes exploration tractable.

\textbf{Anatomy of a Digital Garment: Node Types.} In the Textile IR, we stop treating garments as collections of pixels and start treating them as structured graphs:

\begin{itemize}
\item \emph{PatternPiece}---more than a 2D shape. These nodes encode grain lines (the fabric's ``spine''), seam allowances, and material assignments.
\item \emph{SeamEdge}---the connective tissue. It ensures that if Side A is 10cm, Side B is not magically 12cm---a common ``topological sin'' that AI-generated images commit with abandon.
\item \emph{Dart} and \emph{Notch}---the GPS markers of assembly, conveying how 2D flat-land becomes 3D volume.
\item \emph{MaterialRegion}---zones defining whether the fabric behaves like a silk scarf or a lead apron.
\end{itemize}

\textbf{Constraint Types.} Three categories enforce cross-pillar consistency: (1) \emph{Geometric}---seam length matching, allowance compatibility, grain alignment; (2) \emph{Physics}---stiffness bounds for target drape, tension limits across size range (woven fabrics need different simulation parameters than knits with 50--200\% stretch); (3) \emph{Sustainability}---traceability requirements, end-of-life disassembly constraints.

\textbf{The Verification Ladder: A Seven-Layer Filter for Reality.} Physics simulation is expensive. Real-time LCA is slow. We cannot afford to run a full-scale digital ``fit test'' on a garment that is geometrically impossible. We propose a Verification Ladder---a series of ``cheap'' sanity checks that gate the ``expensive'' engineering.

\textbf{The Sanity Phase (Layers 1--3):}
\begin{itemize}
\item \emph{Layer 1}: Does the code even parse? (instant, deterministic)
\item \emph{Layer 2}: Are the seams the same length? Do they have matching allowances? (milliseconds)
\item \emph{Layer 3}: Is this a manifold? Or have we accidentally designed a Klein bottle that a human cannot actually put on? (milliseconds)
\end{itemize}

\textbf{The Reality Check (Layers 4--5):}
\begin{itemize}
\item \emph{Layer 4}: Does the mesh explode when gravity is turned on? We check for membrane locking---ensuring our digital cotton does not resist bending like cardboard. (seconds, probabilistic)
\item \emph{Layer 5}: We measure the pressure map across diverse body shapes. If a size 3XL gusset shows 15\% excess tension, we flag it before a single thread is cut. (seconds, probabilistic)
\end{itemize}

\textbf{The Industrial Truth (Layers 6--7):}
\begin{itemize}
\item \emph{Layer 6a}: Can we actually fit these patterns on a roll of fabric? (minutes)
\item \emph{Layer 6b} (Robotic Feasibility): Can the target process execute the design? For cut-and-sew, this validates seam accessibility and sewing order. For robotic assembly, this includes \emph{path planning} (reachability, collision avoidance), \emph{grasp stability} (can a gripper hold the panel without slippage?), and thread-path validation. For 3D weaving, it verifies weave-draft compatibility with loom geometry. (minutes)
\item \emph{Layer 7}: The final boss. We query the LCA databases with the exact geometry and material metadata. No more ``guessing'' the carbon footprint; we calculate it based on the actual grams of fibre used. (hours, data-dependent)
\end{itemize}


For typical garments with 15--25 pattern pieces, Layers 1--3 complete in under 100 milliseconds. Geometrically invalid patterns never consume simulation compute. Simulation failures never trigger expensive LCA queries. This hierarchical gating changes computational complexity---not through better algorithms, but by refusing to run expensive operations on inputs that fail cheap checks.

\textbf{The ``What-If'' Engine: Bidirectional Data Flow.} The IR enables feedback loops that distinguish it from unidirectional file conversion. In a legacy workflow, if the drape is bad, you fix the pattern and hope for the best. In the IR, the physics simulation talks back. A simulation failure (e.g., ``excessive wrinkling at armscye'') maps to specific pattern parameters (``increase ease by 2cm''). Swap organic cotton for a hemp blend? The IR instantly updates the LCA dashboard (Carbon: $-2.1$ kg) while warning that higher bending rigidity might require dart repositioning to preserve the silhouette. This is not just a spreadsheet; it is a design-aware ecosystem where every choice has immediate, visible consequences.

The Textile IR exhibits structural isomorphism with scene graphs from robotics and computer vision---garments decompose hierarchically from components to panels to edges, mirroring how buildings decompose into rooms containing objects. This is not mere analogy; it suggests cross-domain transfer is possible. Supplementary data provides detailed visual comparison and discusses how advances in hierarchical scene reasoning may directly apply to garment DSLs.

But why does integration matter epistemically? When systems cannot communicate, uncertainty cannot be propagated. Missing uncertainty quantification undermines the sustainability claims that motivate this entire enterprise.

\section{The Compound Uncertainty Problem}
\label{sec:uncertainty}

When systems cannot communicate, uncertainty compounds invisibly. This has epistemic consequences. Disconnection undermines sustainability claims.

\subsection{Error Propagation Through Disconnected Systems}

Integrated pipelines inherit and amplify uncertainty at each stage. Material characterisation carries measurement uncertainty. Physics simulation introduces numerical error. LCA databases contain impact factor uncertainty. Standard propagation applies: three stages each contributing $\sigma = 15\%$ (illustrative) compound to approximately 26\% \citep{iso_gum_2024}. This matters because current tools present single-point estimates without confidence intervals. A designer sees ``Carbon: 12.3 kg CO$_2$e.'' The honest answer: ``Carbon: 12.3 $\pm$ 3.2 kg CO$_2$e (95\% CI).'' Two materials with nominally different footprints may be statistically indistinguishable.

Conformal prediction \citep{angelopoulos_conformal_2023} offers guaranteed coverage probabilities---assuming exchangeability of calibration and test distributions. This assumption requires validation when material processing or supplier changes occur. Monte Carlo simulation, standard in LCA practice~\citep{he_uncertainty_2025}, could propagate distributions rather than point estimates.

\subsection{Categorical Uncertainty}

Disconnection creates categorical uncertainty: numerically identical metrics can mask experientially different outcomes. Consider ``recycled polyester'' options. Material A (mechanically recycled) and Material B (chemically recycled) show similar carbon footprints, but physical behaviours differ. Mechanical recycling degrades polymer chains. Chemical recycling may introduce contaminants. Without physics connected to LCA, the designer chooses based on carbon alone---then discovers drape problems requiring pattern modifications. The modifications alter the carbon calculation. Full circle.

\subsection{Legal Implications}

The EU Green Claims Directive (2024/825) prohibits ``vague environmental claims without supporting evidence.'' Claims with substantial hidden uncertainty---where confidence intervals span alternative materials---risk violation. The US FTC Green Guides require ``competent and reliable scientific evidence.'' False precision is not competence.

\subsection{Why Integration Is Epistemically Required}

Uncertainty cannot be propagated across systems that do not communicate. Full stop. Integration enables: metadata propagation, correlation tracking, sensitivity analysis, confidence calibration~\citep{marx_uncertainty_2023}.

\subsection{Worked Example: Bidirectional Flow in Practice}
\label{sec:worked-example}

To illustrate operational integration, we trace a complete design-to-validation cycle:

\textbf{Step 1: Material Selection.} A designer selects organic cotton (GOTS-certified, supplier ID: TX-2847) for a flowing summer dress. The Textile IR immediately queries: physics database (bending rigidity B = 0.12 gf$\cdot$cm$^2$/cm, per Kawabata); LCA database (carbon: 8.2 kg CO$_2$e/kg, water: 2,100 L/kg); and traceability layer (supplier data completeness: 78\%, DPP readiness: partial).

\textbf{Step 2: Physics Simulation.} Drape simulation reveals excessive stiffness for the intended silhouette. The diagnostic output: ``Bending rigidity 0.12 exceeds target 0.08 for flowing drape; wrinkle probability at waist seam: 73\%.'' The IR maps this to actionable guidance: ``Consider: (a) material substitution, (b) pattern ease increase +3cm, (c) dart repositioning.''

\textbf{Step 3: Pattern Modification.} The designer chooses option (b): ease increase. The IR propagates: pattern geometry updates (SeamEdge lengths recalculated); marker efficiency decreases (fabric utilisation: 82\% $\rightarrow$ 79\%); LCA updates (material consumption: +3.7\%, carbon: 8.2 $\rightarrow$ 8.5 kg CO$_2$e per garment).

\textbf{Step 4: Validation.} Physics re-simulation confirms acceptable drape. The designer sees a unified dashboard: silhouette preview (satisfactory), cost delta (+\pounds 0.42), carbon delta (+0.3 kg CO$_2$e), and DPP readiness (78\%). The decision trace---material selection, simulation diagnostic, pattern modification, and tradeoff acceptance---becomes provenance for the Digital Product Passport.

This cycle, currently requiring weeks with disconnected tools, could complete in minutes with Textile IR integration. The worked example demonstrates that ``bidirectional flow'' is not abstract: simulation diagnostics map to specific pattern parameters, and pattern changes propagate to quantified sustainability impacts.

\section{Case Illustrations}

\subsection{Material Tradeoff Exploration}

\textbf{Scenario}: A designer develops a jacket considering organic cotton, recycled polyester, hemp-cotton blend.

\textbf{Current workflow}: Request samples. Develop patterns. Wait 2-3 weeks~\citep{parker-strak_challenges_2023}. Evaluate drape physically. Conduct separate LCA~\citep{dhiwar_life_2025}. Receive conflicting feedback: ``Organic cotton drapes well but high water impact. Recycled polyester has good carbon but stiff drape''~\citep{niinimaki_environmental_2020}. Multiple sampling rounds ensue. This is frustrating.

\textbf{Integrated potential}: Select organic cotton---immediately see drape preview, environmental dashboard, cost estimate. Select hemp-cotton---silhouette requires ease adjustment. Pattern modification suggestions appear. Environmental metrics update. Proceed to single validation sample. \emph{Research gap}: No system currently connects material LCA to physics drape to parametric CAD.

\subsection{Inclusive Fit Validation}

Compression leggings targeting XS--3XL require precise fit~\citep{oh_generation_2025}. Linear grade rules create problems at size extremes---discovered only through sampling. Current workflow: often weeks of iteration.

Integration enables: GarmentCode~\citep{korosteleva_garmentcode_2023} generates graded patterns informed by body scan data~\citep{loper_smpl_2015}. Physics validates across representative shapes before any prototype. The system flags: ``3XL hip gusset shows 15\% excess tension.'' Adjustments happen before commitment. \emph{Research gaps}: Physics simulation has not been validated for body shape generalisation, and bidirectional simulation-to-CAD flow remains undeveloped.

\section{The Business Case for Integration}

\subsection{The Industry Reality Check}

Global fibre production: 124 million tonnes (2023). Recirculation: below 0.3\% \citep{gbolarumi_assessment_2022}. The secondhand market grows (\$197B in 2023, projected \$317B by 2027) \citep{thredup_2025_2025}. But here is the paradox. It correlates with increased overall consumption. Consumer segmentation reveals distinct groups---from ``antigreeners'' to ``supergreeners''---each requiring tailored sustainability communication~\citep{martinez-huete_green_2025}.

\subsection{Beyond Compliance}

DPP infrastructure required for compliance may enable more. Authentication for resale. Repair tracking. Warranty management. The integrated framework enables quantifying returns~\citep{reich_beyond_2025, tamm_integrating_2026}.

\subsection{The Upcycling Premium}

Upcycling demonstrates viability. Computational garment reuse~\citep{rags2riches_2025} exemplifies the integration thesis: the system automatically segments existing garments, registers salvageable panels to a pattern library, and generates cutting plans that maximise material recovery---connecting computer vision, constraint satisfaction, and manufacturing validation in a single pipeline. Related work includes PatchUp \citep{mei_patchup_2025}, ScrapReCover \citep{kono_scraprecover_2025}, and Refashion \citep{refashion_2025}. Modular approaches like QUILT \citep{hester_quilt_2025} suggest design paradigms where garments compose of interchangeable modules.

Market evidence: upcycling commands significant premiums~\citep{adiguzel_proud_2021}. Chanel unveiled a recycling platform in 2024 \citep{pan_chanel_2025}. AI-assisted upcycling optimises cut patterns from irregular source materials~\citep{mei_patchup_2025, kono_scraprecover_2025}. The circular economy is not just ethics. It is business.

\section{Research Agenda}

We propose three priorities (Table~\ref{tab:priorities}). Designer studies come first. Deliberately.

\begin{table*}[t]
\centering
\caption{Research agenda: three priorities foregrounding designer understanding.}
\label{tab:priorities}
\tablefontsize
\begin{tabularx}{\textwidth}{@{}clXXXX@{}}
\toprule
\textbf{\#} & \textbf{Priority} & \textbf{Current State} & \textbf{Success Criterion} & \textbf{Practitioner Impact} & \textbf{Communities} \\
\midrule
1 & Designer Studies & Researcher assumptions & Evidence-based guidelines & Tools match designer thinking & HCI, design \\
2 & Pattern Interchange & No material/physics link & Round-trip CAD-sim-LCA & Material changes show drape impact & CG, fashion tech \\
3 & Validation Benchmark & Separate benchmarks & End-to-end evaluation & Confidence virtual matches physical & CG, evaluation \\
\bottomrule
\end{tabularx}
\end{table*}

\textbf{Priority 1: Designer Studies (HCI).} We have spent years building tools for a hypothetical ``average designer'' who probably does not exist. We need rigorous HCI studies to understand how practitioners actually navigate constraints \citep{ryu_effective_2025, melnyk_parametric_2025}. Multi-method approach: ethnographic observation, design probes, participatory workshops, longitudinal deployment. \emph{So what for practice}: determines whether this framework addresses real workflow pain points---or just academic fantasies.

\textbf{Priority 2: Pattern Interchange Formats.} Currently, a file moving from CAD to simulation to LCA loses half its ``soul'' in translation. GarmentCode encodes geometry only; DXF-AAMA loses parametric relationships. Fashion-specific semantics---grain line orientation, seam allowance conventions, ease distribution---require domain extensions to existing CAD interchange standards (ISO 10303 STEP). \emph{So what for practice}: your CLO3D files flow into physics simulation without manual conversion.

\textbf{Priority 3: Validation Benchmark.} The field is currently a Wild West of unverified claims. We need open, end-to-end benchmarks to prove which AI systems actually produce ``sewable'' patterns and which ones are just generating pretty hallucinations. \emph{So what for practice}: a public leaderboard comparing virtual fit against physical ground truth shows which tools actually work.

\subsection{Research Gaps: Ten Open Questions}
\label{sec:research-gaps}

Ten research gaps translate these priorities into testable questions across four clusters.

\textbf{Cluster A: Foundational Infrastructure (RG-1 to RG-3).}
\emph{RG-1: Textile IR Formal Specification.} What formal language bridges CAD, physics, and LCA? Success: round-trip conversion preserving grain direction, seam allowance, and ease with zero semantic loss.
\emph{RG-2: Differentiable Simulation-to-CAD Translation.} Can simulation failures suggest pattern fixes? Isolated acceptance metrics prove insufficient---analogous to how code completion benchmarks miss downstream debugging time. Realistic evaluation requires end-to-end workflow measurement: does total design-to-validation time decrease, or do automated suggestions create cascading modifications that offset initial gains? The Textile IR provides the instrumentation for such holistic assessment.
\emph{RG-3: Metadata-to-Mechanics Validation.} Does \citet{dominguez-elvira_practical_2024}'s metadata-to-mechanics generalise? Success: $R^2 > 0.85$ across fabric families. This ``metadata-to-mechanics'' bridge enables DPP data to directly inform simulation parameters. Service-oriented DPP architectures~\citep{reich_beyond_2025} position passports as feedback loops rather than compliance endpoints; if material characterisation data flows from DPP into physics engines, designers gain real-time drape prediction without manual parameter entry---a speculative but architecturally plausible integration that would collapse the current separation between regulatory documentation and creative tools.

\textbf{Cluster B: Human-Centred Design (RG-4 to RG-5).}
\emph{RG-4: Designer Study Methodology.} Do designers want constraint visibility? Success: evidence-based abstraction guidelines validated with $n \geq 20$ practitioners.
\emph{RG-5: Verification UX.} What interfaces make physics feedback actionable without simulation expertise? Success: $\geq 80\%$ comprehension rate.

\textbf{Cluster C: Data and Validation (RG-6 to RG-8).}
\emph{RG-6: Pattern Interchange Semantics.} Neither GarmentCode nor DXF-AAMA preserves bidirectional links. Success: zero semantic loss in CAD$\leftrightarrow$Simulation$\leftrightarrow$LCA conversion.
\emph{RG-7: Physics-Validated Material Database.} Success: 2,500+ fabrics with KES-validated simulation parameters.
\emph{RG-8: End-to-End Benchmark.} Success: public dataset with virtual-physical tests and performance leaderboard.

\textbf{Cluster D: Physical Characterisation (RG-9 to RG-10).}
\emph{RG-9: Robotic Fashion Metrology.} Fashion needs Scan-to-Garment infrastructure. Success: automated characterisation matching manual KES within 5\% error.
\emph{RG-10: Search Space Characterisation.} What is the size of the fashion topology search space? Hypothesis: $\sim$$10^7$ valid configurations. Success: empirical measurement confirming hierarchical verification filters $>99\%$ of invalid candidates.

\section{Discussion}

\subsection{Implications for Designers}

The framework positions AI as infrastructure, not replacement---designers continue making aesthetic judgments while AI handles engineering translation. Progressive disclosure matters: Level 1 provides traffic-light indicators; Level 2 offers summary metrics; Level 3 exposes full technical detail. LLMs enable hyperpersonalised interfaces \citep{lyu_llmdriven_ux_2025} and malleable UIs~\citep{cao_generative_2025}. A draping expert but LCA novice receives detailed sustainability guidance. A production engineer sees simulation parameters while aesthetics are abstracted.

\textbf{The Agentic Design Stack.} If the Textile IR is the semantic glue, web-based design interfaces (e.g. WebGPU) need an orchestration layer to make this complexity manageable. We propose the \textbf{Model Context Protocol (MCP)} as the ``information rail'' enabling design tools to query heterogeneous data sources---material physics registries, LCA databases, robotic loom capacities---through a standardised interface. Emerging patterns such as \textbf{AG-UI} (Agentic UI) enable bidirectional human-in-the-loop synchronisation: as a designer modifies sleeve curvature, specialised agents can highlight areas where robotic weaving density or seam alignment may become constraints. \textbf{A2UI} (Agent-to-UI) approaches provide declarative rendering of complex feedback---drape confidence intervals, uncertainty maps for carbon footprinting---without overwhelming designers with raw simulation data. Without such UX infrastructure, the framework remains an academic exercise: designers simply will not adopt tools that feel like black boxes.

\subsection{Implications for Industry}

\textbf{New entrant advantage}: Digital-native startups can design their toolchain around a single ``digital thread'' from creative intent to a manufacturable specification. They can instrument data capture at the point where it is created (material tests, fit feedback, machine settings) and feed it back into the representation and verification loop. Incumbents often inherit fragmented PLM/BOM systems, vendor-specific spreadsheets, and tacit knowledge distributed across tiers; improvements requiring coordinated change can look ``architectural'' rather than incremental~\citep{henderson_architectural_1990}.

\textbf{SME adoption barriers}: The framework assumes digitised material libraries, size/fit data, and enough computational and domain expertise to operationalise verification. Many small studios and suppliers have one or two of these, but rarely all three. Practical entry points include shared testing facilities, lightweight ``minimum viable'' metadata schemas, and service models that let SMEs contribute structured data without running large models locally. Without such pathways, benefits accrue primarily to resource-rich enterprises.

\textbf{Supply chain implications}: Integrated workflows compress the time between design decisions and downstream consequences, which is an advantage for reshoring but also exposes where information handoffs are currently lossy. A Textile IR can act as a contract between tiers, making explicit what must be known (e.g., yarn/fabric parameters, allowable tolerances, process constraints) before a design is released to production. This can support tier compaction and microfactories, but only if the representation is backed by reliable measurement and incentives for suppliers to share manufacturability signals; otherwise the pipeline stalls at the first missing field, recreating the familiar ``design says yes, factory says no'' loop~\citep{brozynski_multilevel_2022}.

\textbf{Microfactory vignette}: Consider a reshored knitwear microfactory aiming for a 200--300 unit pilot with a two-week lead time. Halfway through, a locally sourced yarn substitution shifts twist and elasticity; the pattern still ``fits'' in a visual simulator, but the first sewn samples show shoulder creep and seam puckering after wash. The issue is often first noticed in the atelier-like checks---a quick hand-feel and drape test on a swatch under a worklight---well before any model flags it. In a Textile IR workflow, the substitution triggers re-verification against shrinkage and elastic-recovery bounds and updates machine settings (e.g., stitch density, needle choice) before cutting a full batch, turning what is usually late-stage scrap into an early, traceable decision point.

\paragraph{From patternmaker to \emph{garment programmer}.}
A reshored, robotics-enabled pipeline does not eliminate craft; it changes where expertise lives. In an integrated workflow, a senior patternmaker increasingly acts as a \emph{garment programmer}: specifying constraints (ease, balance, grain, tolerances), selecting process assumptions (fixturing, sewing order, allowable stitch classes), and interpreting failures surfaced by verification gates. This role requires ``machine fluency''---not writing low-level control code, but understanding what a robotic cutting cell, a sewing cobot, or a 3D weaving setup can and cannot realise, and how those constraints map back to pattern topology and material choice. Training and tooling should therefore focus on constraint authoring, traceable decision logs, and rapid physical--digital calibration (e.g., swatch tests feeding back into Textile IR parameters), so human judgment stays central while routine checking and documentation become automatic.

\subsection{Beyond Traditional Manufacturing}

\begin{figure}[htbp!]
\centering
\includegraphics[width=\columnwidth]{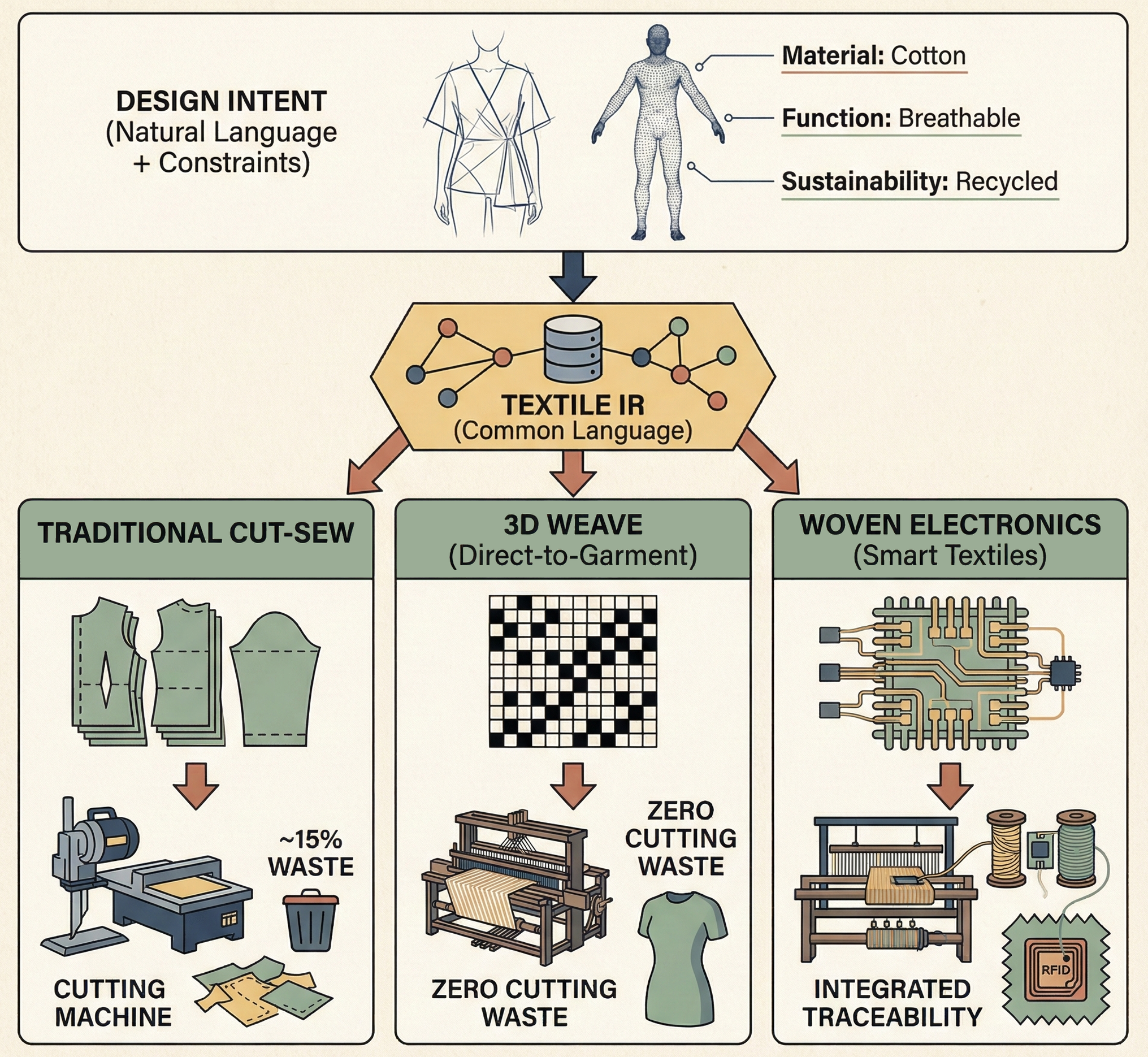}
\caption{\textbf{Program synthesis extends beyond cut-and-sew.} The Textile IR serves as modality-agnostic abstraction: constraints translate to cutting patterns, zero-waste pattern techniques~\citep{lei_pattern_2021}, 3D weave drafts~\citep{perera_evolution_2021, hillaire_all3d_2024}, or woven electronics~\citep{awad_wovencircuits_2025}. Traditional manufacturing incurs ~15\% cutting waste; 3D weaving and zero-waste approaches eliminate this. }
\label{fig:beyond-traditional}
\end{figure}

The program synthesis framework extends beyond cut-and-sew (Figure~\ref{fig:beyond-traditional}). For readers coming from robotics and advanced manufacturing, one helpful lens is \emph{tier compaction}: collapsing handoffs between supply-chain tiers so fewer organisations (and fewer machines) must coordinate to produce a garment. In the Textile Exchange taxonomy, Tier~3 covers yarn-level processing, Tier~2 covers finished material manufacturing (e.g., fabric and trims), and Tier~1 covers finished product assembly~\citep{textileexchange_taxonomy_2024}. Today these tiers are often geographically separated; each boundary introduces batching, minimum order quantities, and metadata loss that slows iteration and makes reshoring hard.

3D weaving is interesting precisely because it can partially collapse the Tier~2$\rightarrow$Tier~1 seam. Instead of weaving ``flat'' fabric and then relying on extensive cut-and-sew labour, recent work demonstrates garments shaped in a single weaving cycle with reduced or eliminated post-loom joining for specific forms~\citep{shi_3dweaving_2024}. Commercial visions such as unspun's Vega\footnote{See \url{https://www.unspun.io/} for their microfactory architecture vision.} describe this as a compact microfactory: yarn-to-garment fabrication co-located with limited finishing, enabling shorter, more responsive local production loops. This is not a magic wand---capital intensity rises, and metrology becomes a bottleneck---but it makes our integration claim sharper: when the physical pipeline is compressed, the representation and verification pipeline must become more explicit.

In this setting, 3D workflows can also reduce physical prototypes~\citep{hillaire_all3d_2024}. Yarn-level homogenisation~\citep{zhang_elasticity_2024} enables accurate physics from material specifications. The catch: computational costs orders of magnitude higher than continuum methods---currently infeasible for real-time iteration. Physical traceability becomes robust when identification is structurally integrated: woven circuits~\citep{awad_wovencircuits_2025} enable RFID throughout garments, not as attached tags but woven in.

\subsection{Implications for Regulation}

Mandating sustainability disclosure without providing tools creates compliance burden without enabling improvement. Requirements encouraging design-stage environmental assessment would accelerate tool development. Regulation shapes markets. Market demand shapes tools.

\subsection{Technological Readiness}

\textbf{What Exists}: Manufacturing-valid pattern generation. Physics-based simulation with zero-shot generalization. Standardised LCA methodology. Multi-criteria decision frameworks.

\textbf{What Is Missing}: Integration infrastructure. No deployed system connects patterns to simulation to impact in real-time loops. The Textile IR provides the architectural specification; implementation requires the research agenda outlined below.

\textbf{What Is Fundamentally Difficult}: The non-differentiability problem. Topology changes create discontinuous landscapes. Gradient-based methods fail. Accurate fabric characterization requires physical testing. This cannot be fully virtualized.

\subsection{Counterarguments}

\textbf{``Fast fashion is the problem.''} Optimizing efficiency could accelerate trend cycles. Fair point. The framework is neutral to use.

\textbf{``Designers may not want invisible engineering.''} Whether designers prefer abstraction or engagement is empirical. Priority 1 addresses this.

\textbf{``Computational frameworks cannot capture tacit knowledge.''} True. \emph{This framework cannot capture making intelligence}---the embodied knowledge of how fabric responds to steam, pressure, and the human hand~\citep{almond_disrupting_2020}. AI handles explicit parameters; the ``soul'' of the garment remains human prerogative.

\subsection{What Would Falsify This Framework?}

Four conditions would undermine our claims. (1) If designer studies show constraint visibility reduces creativity rather than expanding the design space. (2) If compound uncertainty proves negligible (below 5\% rather than our estimated 26\%). (3) If commercial tools achieve integration without formal IR through proprietary mechanisms. (4) If metadata-to-mechanics correlation fails to generalise ($R^2 < 0.5$ on out-of-distribution fabrics). These are testable predictions.

\section{Conclusion}

Four contributions.

First, \textbf{creativity theory}. We argue that engineering constraints become creative resources when AI makes them perceptible, manipulable, immediately consequential~\citep{wu_ai_creativity_2025}. This is a claim. Future empirical work should test it.

Second, \textbf{integration as central challenge}. Generative pattern systems, physics simulation, LCA---each solves important problems. Their value multiplies only when connected. Current tools lack the formal representations for connection. The pieces exist. The glue does not.

Third, the \textbf{Textile IR} as creative infrastructure. Borrowing from compiler theory: textile engineering needs analogous formalisms. Constraint spaces must become legible to creative exploration.

Fourth, \textbf{uncertainty quantification}. When designers see ``12.3 $\pm$ 3.2 kg CO$_2$e (95\% CI)'' rather than false-precision ``12.3 kg CO$_2$e'', they can make informed tradeoffs~\citep{bhatt_uncertainty_2021}. Honest uncertainty enables honest decisions.

For the clothing and textiles research community: pattern engineering, fabric physics, environmental assessment are converging into a single computational substrate.

\subsection{Industry Adoption Pathway}

We expect staged adoption~\citep{mckinsey_state_2025}. Efficiency-first (2025--2027): sampling cost reduction drives early adopters. Compliance-driven (2027--2029): EU DPP requirements~\citep{eu_parliament_dpp_2024} accelerate uptake. Competitive differentiation (2029+): early integrators demonstrate advantage.

\subsection{Call for Collaboration}

\begin{figure*}[htbp!]
\centering
\includegraphics[width=\textwidth]{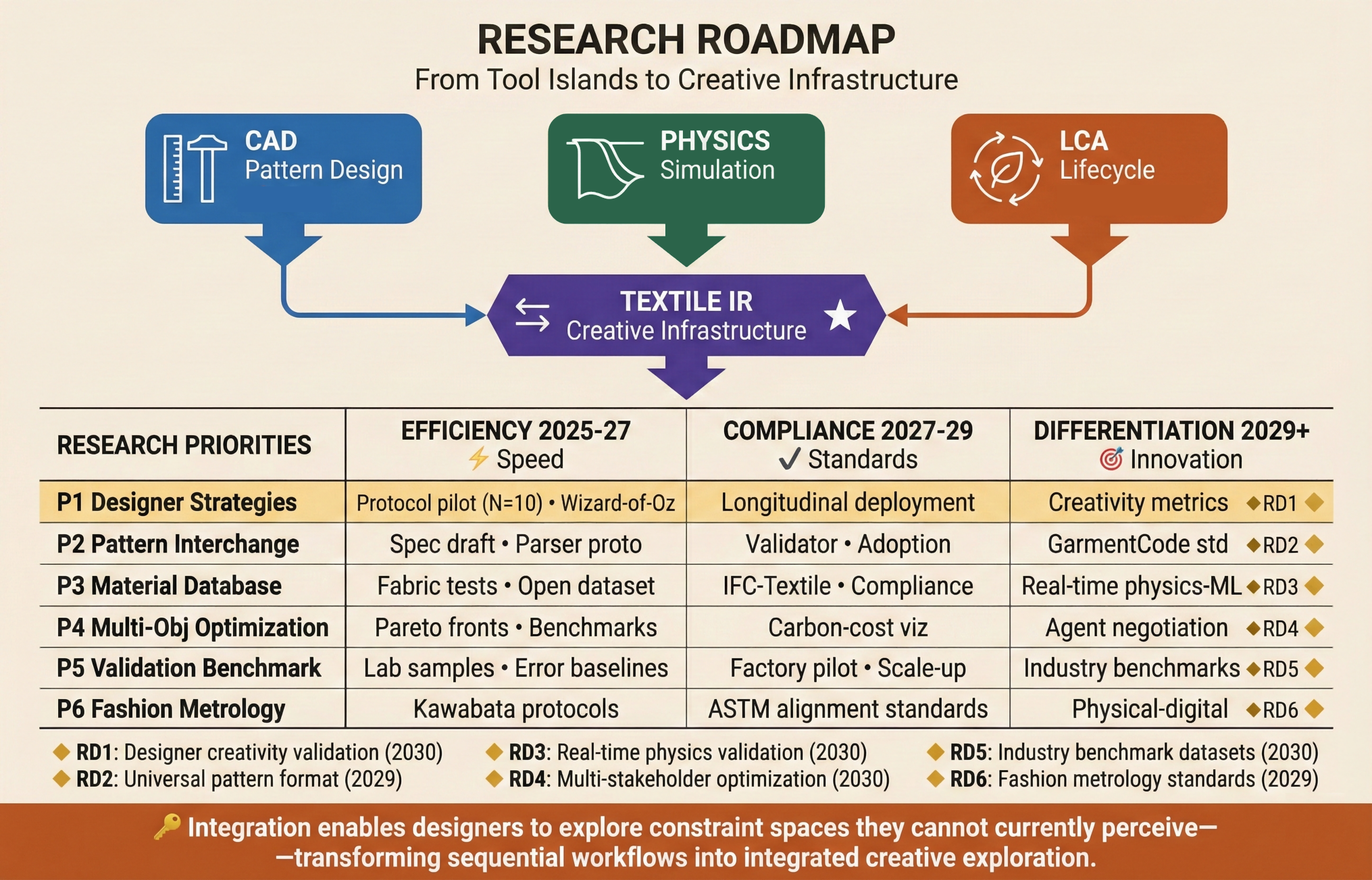}
\caption{\textbf{Research roadmap.} Three foundational pillars (CAD, physics, LCA) converge into the Textile IR. Five research priorities map across adoption phases: efficiency-driven (2025--2027), compliance-driven (2027--2029), and differentiation-driven (2029+). Without integration, circular economy interventions risk rebound backfire~\citep{yerushalmi_rebound_2025}. }
\label{fig:future-research}
\end{figure*}

Our six-priority research agenda (Table~\ref{tab:priorities}) is ambitious. It is also decomposable (Figure~\ref{fig:future-research}). Researchers and practitioners exploring integration build capabilities serving both compliance and competitive strategy.

What changes Monday morning? A designer opens CAD software, selects a material, sees drape preview alongside carbon estimate. Makes informed tradeoff. Commits. The component technologies exist and are validated. The Textile IR specifies how they connect. The question is whether the fashion industry will prioritise integration over continued tool fragmentation.

We invite collaboration across computer graphics, fashion technology, sustainability science, HCI, and design studies.

\subsection*{Acknowledgements}

The authors thank Sabina Weiss for discussions on fashion industry dynamics.

\subsection*{Declaration of Generative AI Usage}

In accordance with Taylor \& Francis policy on AI-assisted authorship, the authors disclose the following uses of generative AI in preparing this manuscript:

\textbf{Writing and editing.} Claude Code (Anthropic Claude Opus 4.5 and Sonnet 4.5) assisted with literature search, proofreading, sentence-level editing, and LaTeX formatting. All AI-generated suggestions were critically reviewed, substantially revised, and validated by the authors. The architectural framework, Textile IR specification, and research priorities represent the authors' original intellectual contribution.

\textbf{LLM-as-judge evaluation.} Multiple LLM models (Claude, Gemini 2.5 Pro, GPT-5.2) were used in an ``LLM Council'' configuration to evaluate draft prose for AI-sounding patterns and suggest humanisation strategies. This meta-use of AI to de-AIify text is itself disclosed.

\textbf{Figures.} Conceptual figure drafts were generated using Gemini 3.0 Pro via Nano Banana Pro (a Figma-based workflow). All figures were reviewed, validated for technical accuracy, and approved by the authors. Figure compositions and scientific content reflect author intent.

\textbf{Data and confidentiality.} No confidential or proprietary data was processed through AI systems. All cited sources are from publicly available literature.

The authors take full responsibility for the accuracy, integrity, and originality of all content in this manuscript.

\subsection*{Data Availability Statement}

This article presents an architectural framework specification based on synthesis of published research. No new empirical data was collected. The Textile IR specification draws on validated systems (GarmentCode, MPMAvatar, PEF methodology) documented in their respective publications. All cited sources are publicly available through standard academic channels.

\subsection*{Ethics Statement}

This research did not involve human participants, animal subjects, or collection of personal data. No ethics approval was required. The article discusses technological frameworks for the fashion industry but does not report experimental results.

\subsection*{Disclosure Statement}

The author(s) declared no potential conflicts of interest with respect to the research, authorship, and/or publication of this article.

\subsection*{Funding}

This research received no specific grant from any funding agency in the public, commercial, or not-for-profit sectors.



\FloatBarrier  



\section*{References}
\addcontentsline{toc}{section}{References}
\begin{thebibliography}{999}

\bibitem[Aakko and Niinimäki(2024)]{aakko_managing_2024} \href{https://doi.org/10.1080/17569370.2024.2338995}{Aakko, M. and Niinimäki, K. (2024)} Managing and Negotiating: Approaches to Quality Work in Clothing and Textiles Companies {\em Fashion Practice 16 391-419}.

\bibitem[Adıgüzel and Donato(2021)]{adiguzel_proud_2021} \href{https://doi.org/10.1016/j.jbusres.2021.03.033}{Adıgüzel, F. and Donato, C. (2021)} Proud to be sustainable: Upcycled versus recycled luxury products {\em Journal of Business Research 130 137-146}.

\bibitem[Aimé et al.(2016)]{aime_vetivoc_2016} \href{https://journals.sagepub.com/action/showAbstract}{Aimé, X. and George, S. and Hornung, J. (2016)} VetiVoc: A modular ontology for the fashion, textile and clothing domain.

\bibitem[Almond(2020)]{almond_disrupting_2020} \href{https://doi.org/10.1080/17569370.2019.1658346}{Almond, K. (2020)} Disrupting the Fashion Archive: The Serendipity of Manufacturing Mistakes {\em Fashion Practice 12 78-101}.

\bibitem[Amershi et al.(2019)]{amershi_guidelines_2019} \href{https://doi.org/10.1145/3290605.3300233}{Amershi, S., Weld, D., Vorvoreanu, M., Fourney, A., Nushi, B., Collisson, P., Suh, J., Iqbal, S., Bennett, P.N., Inkpen, K. et al. (2019)} Guidelines for {Human-AI} Interaction {\em Proceedings of the 2019 CHI Conference on Human Factors in Computing Systems 1-13}.

\bibitem[Angelopoulos and Bates(2023)]{angelopoulos_conformal_2023} \href{https://doi.org/10.1561/2200000101}{Angelopoulos, A.N. and Bates, S. (2023)} Conformal Prediction: A Gentle Introduction {\em Foundations and Trends in Machine Learning 16 494-591}.

\bibitem[Awad et al.(2025)]{awad_wovencircuits_2025} \href{https://doi.org/10.1145/3715336.3735696}{Awad, A. and Ibrahim, S. and Nabil, S. (2025)} WovenCircuits: A {3-Step} Fabrication Process for Weaving Electric Circuit Layouts in Everyday Artefacts {\em Proceedings of the 2025 ACM Designing Interactive Systems Conference 2428–2444}.

\bibitem[Balloni et al.(2025)]{balloni_neural_2025} \href{https://doi.org/10.1016/j.engappai.2025.110773}{Balloni, E., Stacchio, L., Mancini, A., Frontoni, E., Zingaretti, P. and Paolanti, M. (2025)} A Neural Rendering system for fashion design process {\em Engineering Applications of Artificial Intelligence 153 110773}.

\bibitem[Ban et al.(2025)]{ban_sagsgnn_2025} \href{https://doi.org/10.1016/j.cag.2025.104216}{Ban, K., Huai, Y., Nie, X., Meng, Q. and Xu, H. (2025)} SAGS-GNN: Graph Neural Network for self-collision and anisotropy in dynamic garment simulation {\em Computers \& Graphics 128 104216}.

\bibitem[Baraff and Witkin(1998)]{baraff_large_1998} \href{https://doi.org/10.1145/280814.280821}{Baraff, D. and Witkin, A. (1998)} Large steps in cloth simulation {\em Proceedings of the 25th annual conference on Computer graphics and interactive techniques  - SIGGRAPH '98 43-54}.

\bibitem[Bertola and Teunissen(2018)]{bertola_fashion_2018} \href{https://doi.org/10.1108/RJTA-03-2018-0023}{Bertola, P. and Teunissen, J. (2018)} Fashion 4.0. Innovating fashion industry through digital transformation {\em Research Journal of Textile and Apparel 22 352-369}.

\bibitem[Bhatt et al.(2021)]{bhatt_uncertainty_2021} \href{https://dl.acm.org/doi/10.1145/3461702.3462571}{Bhatt, U., Antorán, J., Zhang, Y., Liao, Q.V., Sattigeri, P., Fogliato, R., Melançon, G., Krishnan, R., Stanley, J., Tickoo, O. et al. (2021)} Uncertainty as a Form of Transparency: Measuring, Communicating, and Using Uncertainty {\em Proceedings of the 2021 AAAI/ACM Conference on AI, Ethics, and Society 401–413}.

\bibitem[Bian et al.(2025)]{bian_chatgarment_2025} \href{https://doi.org/10.48550/arXiv.2412.17811}{Bian, S., Xu, C., Xiu, Y., Grigorev, A., Liu, Z., Lu, C., Black, M.J. and Feng, Y. (2025)} ChatGarment: Garment Estimation, Generation and Editing via Large Language Models {\em arXiv:2412.17811}.

\bibitem[Bocken et al.(2022)]{bocken_sufficiency_2022} \href{https://www.frontiersin.org/journals/sustainability/articles/10.3389/frsus.2022.899289/full}{Bocken, N.M.P. and Niessen, L. and Short, S.W. (2022)} The {Sufficiency-Based} Circular Economy—An Analysis of 150 Companies.

\bibitem[Brozynski and Leibowicz(2022)]{brozynski_multilevel_2022} \href{https://doi.org/10.1016/j.ejor.2021.10.026}{Brozynski, M.T. and Leibowicz, B.D. (2022)} A multi-level optimization model of infrastructure-dependent technology adoption: Overcoming the chicken-and-egg problem {\em European Journal of Operational Research 300 755-770}.

\bibitem[Cao et al.(2025)]{cao_generative_2025} \href{https://doi.org/10.48550/arXiv.2503.04084}{Cao, Y. and Jiang, P. and Xia, H. (2025)} Generative and Malleable User Interfaces with Generative and Evolving {Task-Driven} Data Model {\em arXiv:2503.04084}.

\bibitem[Chaudhary et al.(2020)]{chaudhary_maximizing_2020} \href{https://doi.org/10.1177/1847979020975528}{Chaudhary, S. and Kumar, P. and Johri, P. (2020)} Maximizing performance of apparel manufacturing industry through {CAD} adoption {\em International Journal of Engineering Business Management 12 184797902097552}.

\bibitem[Chen(2025)]{chen_automating_2025} \href{https://doi.org/10.17918/00011002}{Chen, J. (2025)} Automating garment pattern making with {AI}: evaluating the performance and practical utility of fine-tuned large language models in fashion production {\em Master's Thesis, University of the Arts London}.

\bibitem[Chen et al.(2025)]{chen_learning_2025} \href{https://doi.org/10.48550/arXiv.2507.21288}{Chen, G., Suri, S., Wu, Y., Voulga, E., Levin, D.I.W. and Pai, D.K. (2025)} Learning Simulatable Models of Cloth with {Spatially-varying} Constitutive Properties {\em arXiv:2507.21288}.

\bibitem[Chen et al.(2022)]{chen_structurepreserving_2022} \href{https://doi.org/10.48550/arXiv.2211.06701}{Chen, X., Wang, G., Zhu, D., Liang, X., Torr, P.H.S. and Lin, L. (2022)} Structure-Preserving {3D} Garment Modeling with Neural Sewing Machines {\em arXiv:2211.06701}.

\bibitem[Chollet(2019)]{chollet_measure_2019} \href{https://arxiv.org/abs/1911.01547}{Chollet, F. (2019)} On the Measure of Intelligence {\em arXiv:1911.01547}.

\bibitem[Chow et al.(2005)]{chow_design_2005} \href{https://doi.org/10.1016/j.eswa.2005.04.001}{Chow, H., Choy, K., Lee, W. and Chan, F. (2005)} Design of a knowledge-based logistics strategy system {\em Expert Systems with Applications 29 272-290}.

\bibitem[Dhiwar and Bedarkar(2025)]{dhiwar_life_2025} \href{https://doi.org/10.1007/s43621-025-02050-7}{Dhiwar, K. and Bedarkar, M. (2025)} Life cycle assessment in fashion industry: a systematic review {\em Discover Sustainability 6}.

\bibitem[Dominguez-Elvira et al.(2024)]{dominguez-elvira_practical_2024} \href{https://doi.org/10.1111/cgf.15029}{Dominguez-Elvira, H., Nicas, A., Cirio, G., Rodriguez, A. and Garces, E. (2024)} Practical Methods to Estimate Fabric Mechanics from Metadata {\em Computer Graphics Forum 43 e15029}.

\bibitem[Dove et al.(2017)]{dove_design_2017} \href{https://doi.org/10.1145/3025453.3025739}{Dove, G., Halskov, K., Forlizzi, J. and Zimmerman, J. (2017)} UX Design Innovation {\em Proceedings of the 2017 CHI Conference on Human Factors in Computing Systems 278-288}.

\bibitem[Dumoulin et al.(2025)]{zhang_dgarment_2025} \href{https://doi.org/10.48550/arXiv.2504.03468}{Dumoulin, A., Boukhayma, A., Boissieux, L., Damodaran, B.B., Hellier, P. and Wuhrer, S. (2025)} D-Garment: {Physics-Conditioned} Latent Diffusion for Dynamic Garment Deformations {\em arXiv:2504.03468}.

\bibitem[European Union(2024)]{eu_parliament_dpp_2024} \href{http://data.europa.eu/eli/reg/2024/1781/oj}{European Union (2024)} Regulation ({EU}) 2024/1781 Establishing a Framework for Ecodesign Requirements for Sustainable Products ({ESPR}).

\bibitem[Faruqi et al.(2024)]{faruqi_shaping_2024} \href{https://doi.org/10.48550/arXiv.2404.10142}{Faruqi, F., Tian, Y., Phadnis, V., Jampani, V. and Mueller, S. (2024)} Shaping Realities: Enhancing {3D} Generative {AI} with Fabrication Constraints {\em arXiv:2404.10142}.

\bibitem[Fawzi et al.(2022)]{fawzi_discovering_2022} \href{https://doi.org/10.1038/s41586-022-05172-4}{Fawzi, A., Balog, M., Huang, A., Hubert, T., Romera-Paredes, B., Barekatain, M., Novikov, A., R. Ruiz, F.J., Schrittwieser, J., Swirszcz, G. et al. (2022)} Discovering faster matrix multiplication algorithms with reinforcement learning {\em Nature 610 47-53}.

\bibitem[Gbolarumi and Wong(2022)]{gbolarumi_assessment_2022} Gbolarumi, F.T. and Wong, K.Y. (2022) Assessment of Environmental Performance Criteria in Textile Industry Using the {Best-Worst} Method {\em Advances in Best-Worst Method 160-174}.

\bibitem[Govindan et al.(2015)]{govindan_multi_2015} \href{https://doi.org/10.1016/j.jclepro.2013.06.046}{Govindan, K., Rajendran, S., Sarkis, J. and Murugesan, P. (2015)} Multi criteria decision making approaches for green supplier evaluation and selection: a literature review {\em Journal of Cleaner Production 98 66-83}.

\bibitem[Guo and Istook(2023)]{guo_evaluation_2023} \href{https://doi.org/10.1080/17569370.2021.1987647}{Guo, S. and Istook, C.L. (2023)} Evaluation of {2D} {CAD} Technology for Garments Customized for Body Shape {\em Fashion Practice 15 136-162}.

\bibitem[Gutin et al.(2015)]{gutin_interdiction_2015} \href{https://doi.org/10.1287/mnsc.2014.1973}{Gutin, E. and Kuhn, D. and Wiesemann, W. (2015)} Interdiction Games on Markovian {PERT} Networks {\em Management Science 61 999-1017}.

\bibitem[He et al.(2024)]{he_dresscode_2024} \href{https://doi.org/10.1145/3658147}{He, K., Yao, K., Zhang, Q., Yu, J., Liu, L. and Xu, L. (2024)} DressCode: Autoregressively Sewing and Generating Garments from Text Guidance {\em ACM Trans. Graph. 43 72:1–72:13}.

\bibitem[He et al.(2025)]{he_uncertainty_2025} \href{https://doi.org/10.1016/j.jclepro.2025.144885}{He, Q. and Wu, X. and Ding, X. (2025)} Uncertainty in the Carbon Footprint accounting and evaluation of textile and apparel products: A systematic review {\em Journal of Cleaner Production 492 144885}.

\bibitem[Henderson and Clark(1990)]{henderson_architectural_1990} \href{https://doi.org/10.2307/2393549}{Henderson, R.M. and Clark, K.B. (1990)} Architectural Innovation: The Reconfiguration of Existing Product Technologies and the Failure of Established Firms {\em Administrative Science Quarterly 35 9}.

\bibitem[Hester et al.(2025)]{hester_quilt_2025} \href{https://dl.acm.org/doi/10.1145/3746059.3747608}{Hester, J. and Law, S. and Hofmann, M. (2025)} QUILT: Supporting Modular Design of {Machine-Knitting} Programs {\em Proceedings of the 38th Annual ACM Symposium on User Interface Software and Technology 1–26}.

\bibitem[Hillaire and Baytar(2024)]{hillaire_all3d_2024} \href{https://doi.org/10.1177/15589250241252634}{Hillaire, J.M. and Baytar, F. (2024)} All-3D apparel development: Establishing the rules to enable {3D} weaving from {3D} digital garments {\em Journal of Engineered Fibers and Fabrics 19 15589250241252634}.

\bibitem[Islam et al.(2024)]{islam_deep_2024} \href{https://doi.org/10.1109/ACCESS.2024.3368612}{Islam, T., Miron, A., Liu, X. and Li, Y. (2024)} Deep Learning in Virtual {Try-On}: A Comprehensive Survey {\em IEEE Access 12 29475-29502}.

\bibitem[Iso(2024)]{iso_gum_2024} \href{https://www.iso.org/obp/ui/en/#iso:std:iso-iec:guide:98:-1:ed-2:v1:en}{Iso (2024)} ISO/{IEC} Guide 98-1:2024(en), Guide to the expression of uncertainty in measurement.

\bibitem[Jiang et al.(2025)]{jiang_avatarvton:_2025} \href{https://doi.org/10.48550/arXiv.2510.04822}{Jiang, Z., Gao, J., He, S., Li, X., Zheng, Y., Yang, Z., Dong, J. and Du, Y. (2025)} AvatarVTON: {4D} Virtual {Try-On} for Animatable Avatars {\em arXiv:2510.04822}.

\bibitem[Kerbl et al.(2023)]{kerbl_3d_2023} \href{https://doi.org/10.1145/3592433}{Kerbl, B., Kopanas, G., Leimkuehler, T. and Drettakis, G. (2023)} 3D Gaussian Splatting for {Real-Time} Radiance Field Rendering.

\bibitem[Kim(2020)]{kim_finite_2020} \href{https://doi.org/10.1111/cgf.14111}{Kim, T. (2020)} A Finite Element Formulation of Baraff‐Witkin Cloth {\em Computer Graphics Forum 39 171-179}.

\bibitem[Kodnongbua et al.(2025)]{kodnongbua_design_2025} \href{https://doi.org/10.1145/3757377.3764004}{Kodnongbua, M., Zhang, Z., Sharp, N. and Schulz, A. (2025)} Design for Descent: What Makes a Shape Grammar Easy to Optimize? {\em Proceedings of the SIGGRAPH Asia 2025 Conference Papers 1–11}.

\bibitem[Kono et al.(2025)]{kono_scraprecover_2025} \href{https://doi.org/10.1145/3745778.3766653}{Kono, M., Larsson, M., Shen, I. and Igarashi, T. (2025)} ScrapReCover: An Interactive Optimization System for Freeform Patchwork Layouts {\em Proceedings of the ACM Symposium on Computational Fabrication 1–15}.

\bibitem[Korosteleva and Sorkine-Hornung(2023)]{korosteleva_garmentcode_2023} \href{https://doi.org/10.1145/3618351}{Korosteleva, M. and Sorkine-Hornung, O. (2023)} GarmentCode: Programming Parametric Sewing Patterns {\em ACM Trans. Graph. 42 199:1–199:15}.

\bibitem[Korosteleva et al.(2025)]{korosteleva_garmentcodedata_2024} Korosteleva, M., Kesdogan, T.L., Kemper, F., Wenninger, S., Koller, J., Zhang, Y., Botsch, M. and Sorkine-Hornung, O. (2025) GarmentCodeData: A Dataset of {3D} {Made-to-Measure} Garments with Sewing Patterns {\em Computer Vision – ECCV 2024 110-127}.

\bibitem[Korosteleva and Lee(2022)]{korosteleva_neuraltailor_2022} \href{https://doi.org/10.1145/3528223.3530179}{Korosteleva, M. and Lee, S. (2022)} NeuralTailor: reconstructing sewing pattern structures from {3D} point clouds of garments {\em ACM Trans. Graph. 41 158:1–158:16}.

\bibitem[Langley et al.(2023)]{langley_orchestrating_2023} \href{https://doi.org/10.1016/j.jbusres.2023.114259}{Langley, D.J., Rosca, E., Angelopoulos, M., Kamminga, O. and Hooijer, C. (2023)} Orchestrating a smart circular economy: Guiding principles for digital product passports {\em Journal of Business Research 169 114259}.

\bibitem[Leal Filho et al.(2022)]{filho_overview_2022} \href{https://doi.org/10.3389/fenvs.2022.973102}{Leal Filho, W., Perry, P., Heim, H., Dinis, M.A.P., Moda, H., Ebhuoma, E. and Paço, A. (2022)} An overview of the contribution of the textiles sector to climate change {\em Frontiers in Environmental Science 10}.

\bibitem[Lee et al.(2025)]{lee_mpmavatar_2025} \href{https://arxiv.org/abs/2510.01619}{Lee, C. and Lee, J. and Kim, T. (2025)} MPMAvatar: Learning {3D} Gaussian Avatars with Accurate and Robust {Physics-Based} Dynamics {\em arXiv:2510.01619}.

\bibitem[Lei and Li(2021)]{lei_pattern_2021} \href{https://doi.org/10.1080/17569370.2021.1982503}{Lei, G. and Li, X. (2021)} A Pattern Making Approach to Improving {Zero-Waste} Fashion Design {\em Fashion Practice 13 443-463}.

\bibitem[Li et al.(2025)]{li_dress-1-to-3_2025} \href{https://doi.org/10.1145/3731177}{Li, X., Yu, C., Du, W., Jiang, Y., Xie, T., Chen, Y., Yang, Y. and Jiang, C. (2025)} Dress-1-to-3: Single Image to {Simulation-Ready} {3D} Outfit with Diffusion Prior and Differentiable Physics {\em ACM Trans. Graph. 44 71:1–71:16}.

\bibitem[Li et al.(2025)]{li_ragdiffusion_2025} \href{https://doi.org/10.48550/arXiv.2411.19528}{Li, Y., Tan, X., Shang, W., Wu, Y., Wang, J., Chen, X., Zhang, Y., Lin, R. and Ni, B. (2025)} RAGDiffusion: Faithful Cloth Generation via External Knowledge Assimilation {\em arXiv:2411.19528}.

\bibitem[Li et al.(2025)]{yang_dso_2025} \href{https://doi.org/10.48550/arXiv.2503.22677}{Li, R., Zheng, C., Rupprecht, C. and Vedaldi, A. (2025)} DSO: Aligning {3D} Generators with Simulation Feedback for Physical Soundness {\em arXiv:2503.22677}.

\bibitem[Lin et al.(2025)]{refashion_2025} \href{https://dl.acm.org/doi/10.1145/3746059.3747632}{Lin, R. and Lukáč, M. and Leake, M. (2025)} Refashion: Reconfigurable Garments via Modular Design {\em Proceedings of the 38th Annual ACM Symposium on User Interface Software and Technology 1–18}.

\bibitem[Liu et al.(2018)]{liu_interactive_2018} \href{https://doi.org/10.1016/j.cad.2018.07.003}{Liu, K., Zeng, X., Bruniaux, P., Tao, X., Yao, X., Li, V. and Wang, J. (2018)} 3D interactive garment pattern-making technology {\em Computer-Aided Design 104 113-124}.

\bibitem[Loper et al.(2015)]{loper_smpl_2015} \href{https://doi.org/10.1145/2816795.2818013}{Loper, M., Mahmood, N., Romero, J., Pons-Moll, G. and Black, M.J. (2015)} SMPL {\em ACM Transactions on Graphics 34 1-16}.

\bibitem[Lu et al.(2025)]{lyu_llmdriven_ux_2025} \href{https://doi.org/10.1145/3706599.3719729}{Lu, Y., Yao, B., Gu, H., Huang, J., Wang, Z.J., Li, Y., Gesi, J., He, Q., Li, T.J. and Wang, D. (2025)} UXAgent: An {LLM} {Agent-Based} Usability Testing Framework for Web Design {\em Proceedings of the Extended Abstracts of the CHI Conference on Human Factors in Computing Systems 1–12}.

\bibitem[Ma et al.(2025)]{ma_bim_lca_2025} \href{https://www.sciencedirect.com/science/article/pii/S036013232501159X}{Ma, X., Huang, M., Chen, X., Bai, Y. and Zhang, Q. (2025)} BIM-integrated {LCA} framework for prefabricated buildings with automated benchmarking and visual decision support.

\bibitem[Martinez-Huete and Aramendia-Muneta(2025)]{martinez-huete_green_2025} \href{https://doi.org/10.1080/17569370.2024.2426647}{Martinez-Huete, L. and Aramendia-Muneta, M.E. (2025)} Green Products in the Fashion Industry: Consumer Segmentation to Develop Communication Campaigns {\em Fashion Practice 17 105-132}.

\bibitem[Marx et al.(2023)]{marx_uncertainty_2023} \href{https://proceedings.mlr.press/v206/marx23a.html}{Marx, C., Park, Y., Hasson, H., Wang, Y., Ermon, S. and Huan, L. (2023)} But Are You Sure? An {Uncertainty-Aware} Perspective on Explainable {AI} {\em Proceedings of The 26th International Conference on Artificial Intelligence and Statistics 7375-7391}.

\bibitem[McKinsey \& Company(2025)]{mckinsey_state_2025} \href{https://www.mckinsey.com/industries/retail/our-insights/state-of-fashion}{McKinsey \& Company (2025)} The State of Fashion 2026: When the rules change | McKinsey.

\bibitem[Mei et al.(2025)]{mei_patchup_2025} \href{https://dl.acm.org/doi/10.1145/3745778.3766667}{Mei, Y., Que, L., Leake, M. and Schulz, A. (2025)} PatchUp: Interactive Patchwork Design for Scrap Fabric Upcycling {\em Proceedings of the ACM Symposium on Computational Fabrication 1–17}.

\bibitem[Melnyk(2025)]{melnyk_parametric_2025} \href{https://doi.org/10.1080/14626268.2025.2514120}{Melnyk, V. (2025)} Parametric pattern design for manually knitted textile panels {\em Digital Creativity 36 167-177}.

\bibitem[Mildenhall et al.(2020)]{mildenhall_nerf_2020} \href{https://doi.org/10.48550/arXiv.2003.08934}{Mildenhall, B., Srinivasan, P.P., Tancik, M., Barron, J.T., Ramamoorthi, R. and Ng, R. (2020)} NeRF: Representing Scenes as Neural Radiance Fields for View Synthesis {\em arXiv:2003.08934}.

\bibitem[Mizrachi and Sharon(2025)]{nature_secondhand_2025} \href{https://doi.org/10.1038/s41598-025-19089-1}{Mizrachi, M.P. and Sharon, O. (2025)} Secondhand fashion consumers exhibit fast fashion behaviors despite sustainability narratives {\em Scientific Reports 15 34968}.

\bibitem[Murthy et al.(2020)]{murthy_gradsim_2020} \href{https://openreview.net/forum?id=c_E8kFWfhp0}{Murthy, J.K., Macklin, M., Golemo, F., Voleti, V., Petrini, L., Weiss, M., Considine, B., Parent-Lévesque, J., Xie, K., Erleben, K. et al. (2020)} gradSim: Differentiable simulation for system identification and visuomotor control.

\bibitem[Nakayama et al.(2025)]{nakayama_aipparel_2025} \href{https://openaccess.thecvf.com/content/CVPR2025/html/Nakayama_AIpparel_A_Multimodal_Foundation_Model_for_Digital_Garments_CVPR_2025_paper.html}{Nakayama, K., Ackermann, J., Kesdogan, T.L., Zheng, Y., Korosteleva, M., Sorkine-Hornung, O., Guibas, L.J., Yang, G. and Wetzstein, G. (2025)} AIpparel: A Multimodal Foundation Model for Digital Garments.

\bibitem[Navarrete et al.(2021)]{navarrete_flue_2021} \href{https://doi.org/10.1016/j.jclepro.2020.124646}{Navarrete, I., Vargas, F., Martinez, P., Paul, A. and Lopez, M. (2021)} Flue gas desulfurization ({FGD}) fly ash as a sustainable, safe alternative for cement-based materials {\em Journal of Cleaner Production 283 124646}.

\bibitem[Niinimäki et al.(2020)]{niinimaki_environmental_2020} \href{https://doi.org/10.1038/s43017-020-0039-9}{Niinimäki, K., Peters, G., Dahlbo, H., Perry, P., Rissanen, T. and Gwilt, A. (2020)} The environmental price of fast fashion {\em Nature Reviews Earth \& Environment 1 189-200}.

\bibitem[Nisa et al.(2025)]{nisa_systematic_2025} \href{https://doi.org/10.3390/app15105691}{Nisa, H., Van Amber, R., English, J., Islam, S., McCorkill, G. and Alavi, A. (2025)} A Systematic Review of Reimagining Fashion and Textiles Sustainability with {AI}: A Circular Economy Approach {\em Applied Sciences 15 5691}.

\bibitem[Oh and Kim(2025)]{oh_generation_2025} \href{https://doi.org/10.1177/0887302X251340652}{Oh, J. and Kim, S. (2025)} Generation of {Body-Fit} Garment Patterns Using a Landmark Matching Algorithm {\em Clothing and Textiles Research Journal}.

\bibitem[Palomo-Lovinski(2024)]{palomo-lovinski_missed_2024} \href{https://doi.org/10.1080/17569370.2024.2312925}{Palomo-Lovinski, N. (2024)} Missed Opportunities: Fashion Fabric Sourcing Professionals’ Use of the {MSI} in the Higg Index {\em Fashion Practice 16 373-390}.

\bibitem[Pan(2025)]{pan_chanel_2025} \href{https://www.vogue.com/article/chanel-unveils-new-recycling-platform}{Pan, Y. (2025)} Chanel unveils new recycling platform.

\bibitem[Parker-Strak et al.(2023)]{parker-strak_challenges_2023} \href{https://doi.org/10.1080/17569370.2023.2247907}{Parker-Strak, R. and Doyle, S. and Studd, R. (2023)} Challenges and Changes to the Product Development Process for Fashion Omnichannel Retailers {\em Fashion Practice 16 81-107}.

\bibitem[Perera et al.(2021)]{perera_evolution_2021} \href{https://doi.org/10.1186/s40691-020-00240-7}{Perera, Y.S., Muwanwella, R.M.H.W., Fernando, P.R., Fernando, S.K. and Jayawardana, T.S.S. (2021)} Evolution of {3D} weaving and {3D} woven fabric structures {\em Fashion and Textiles 8 11}.

\bibitem[Pourjafarian et al.(2025)]{pourjafarian_odp-based_2025} \href{https://doi.org/10.1016/j.procir.2024.12.125}{Pourjafarian, M., Plociennik, C., Bergweiler, S., Moarefvand, N., Brozeit, J., Rezapour, M. and Ruskowski, M. (2025)} An {ODP-based} Ontology for the Digital Product Passport {\em Procedia CIRP 135 930-935}.

\bibitem[Qi et al.(2025)]{rags2riches_2025} \href{https://dl.acm.org/doi/10.1145/3721238.3730703}{Qi, A., Pietroni, N., Korosteleva, M., Sorkine-Hornung, O. and Bousseau, A. (2025)} Rags2Riches: Computational Garment Reuse {\em Proceedings of the Special Interest Group on Computer Graphics and Interactive Techniques Conference Conference Papers 1–11}.

\bibitem[Ramírez-Escamilla et al.(2024)]{ramirez-escamilla_advancing_2024} \href{https://doi.org/10.3390/recycling9050095}{Ramírez-Escamilla, H.G., Martínez-Rodríguez, M.C., Padilla-Rivera, A., Domínguez-Solís, D. and Campos-Villegas, L.E. (2024)} Advancing Toward Sustainability: A Systematic Review of Circular Economy Strategies in the Textile Industry {\em Recycling 9 95}.

\bibitem[Reich et al.(2025)]{reich_beyond_2025} \href{https://doi.org/10.1016/j.procir.2024.12.120}{Reich, R.H. and Alaerts, L. and Acker, K.V. (2025)} Beyond compliance: Designing value-creating, service-oriented, and integrated product passports that contribute to the circular economy {\em Procedia CIRP 135 997-1002}.

\bibitem[Richardy et al.(2025)]{richardy_wicked_2025} \href{https://doi.org/10.54337/plate2025-10392}{Richardy, J. and Skjold, E. and Skødt, T. (2025)} The Wicked Problems of Durability: Rebound Effects and Textile Illiteracy in Circular Policy {\em Proceedings of the 6th Product Lifetimes and the Environment Conference (PLATE2025)}.

\bibitem[Rizzi and Bertola(2025)]{rizzi_generative_2025} \href{https://doi.org/10.3389/ejcmp.2025.13875}{Rizzi, G. and Bertola, P. (2025)} Exploring the generative {AI} potential in the fashion design process: an experimental experience on the collaboration between fashion design practitioners and generative {AI} tools.

\bibitem[Romera-Paredes et al.(2024)]{romera-paredes_mathematical_2024} \href{https://doi.org/10.1038/s41586-023-06924-6}{Romera-Paredes, B., Barekatain, M., Novikov, A., Balog, M., Kumar, M.P., Dupont, E., Ruiz, F.J.R., Ellenberg, J.S., Wang, P., Fawzi, O. et al. (2024)} Mathematical discoveries from program search with large language models.

\bibitem[Ryu and Lee(2025)]{ryu_effective_2025} \href{https://doi.org/10.1177/0887302X251348003}{Ryu, C. and Lee, Y.K. (2025)} Effective Fashion Design Collection Implementation with Generative {AI}: {ChatGPT} and {DALL-E} {\em Clothing and Textiles Research Journal}.

\bibitem[Santos et al.(2025)]{santos_circularity_2025} \href{https://doi.org/10.1002/csr.70181}{Santos, T.V.D., Alvez, S.M.D., Sehnem, S. and Julkovski, D.J. (2025)} The Circularity Paradox: The Rebound Effect, Governance, and the Limits of Corporate Sustainability.

\bibitem[Selkee(2025)]{selkee_ai-driven_2025} \href{https://lauda.ulapland.fi/bitstream/handle/10024/66758/Selkee_Charlotta.pdf?sequence=1}{Selkee, C. (2025)} AI-driven innovation in fashion: Enhancing the design process through technology.

\bibitem[Shi et al.(2024)]{shi_3dweaving_2024} \href{https://www.cell.com/iscience/abstract/S2589-0042(24)01540-2}{Shi, Y., Taylor, L.W., Kulessa, A., Cheung, V. and Sayem, A.S.M. (2024)} Re-engineer apparel manufacturing processes with {3D} weaving technology for efficient single-step garment production.

\bibitem[Shi et al.(2025)]{shi_genai_fashion_2025} \href{https://doi.org/10.1145/3718098}{Shi, W. and Wong, W. and Zou, X. (2025)} Generative {AI} in Fashion: Overview {\em ACM Trans. Intell. Syst. Technol. 16 74:1–74:73}.

\bibitem[Song et al.(2023)]{song_imagebased_2023} \href{https://arxiv.org/abs/2311.04811}{Song, D., Zhang, X., Zhou, J., Nie, W., Tong, R., Kankanhalli, M. and Liu, A. (2023)} Image-Based Virtual {Try-On}: A Survey {\em arXiv:2311.04811}.

\bibitem[Stuyck(2018)]{stuyck_cloth_2018} \href{https://doi.org/10.1007/978-3-031-02597-6}{Stuyck, T. (2018)} Cloth Simulation for Computer Graphics {\em Synthesis Lectures on Visual Computing: Computer Graphics, Animation, Computational Photography and Imaging}.

\bibitem[Sun et al.(2025)]{sun_deep_2025} \href{https://doi.org/10.1177/00405175251335188}{Sun, Y., Hao, Z., Wang, Z., Jin, J., Ye, Q. and Lyu, Y. (2025)} Deep learning for {3D} garment generation: A review {\em Textile Research Journal 00405175251335188}.

\bibitem[Sun et al.(2025)]{tailor_2025} \href{https://doi.org/10.48550/arXiv.2503.12052}{Sun, Z., Wen, Y., Lin, M., Fang, H., Ye, S., Lv, T. and Liu, Y. (2025)} Tailor: An Integrated {Text-Driven} {CG-Ready} Human and Garment Generation System {\em arXiv:2503.12052}.

\bibitem[Tamm et al.(2026)]{tamm_integrating_2026} \href{https://doi.org/10.1007/978-3-032-11976-6_23}{Tamm, M., Draheim, D., Tammet, T. and Pappel, I. (2026)} Integrating Digital Product Passports in {E-Commerce}: An Architectural Framework for Sustainability and {AI-Driven} Value {\em Information Integration and Web Intelligence 315-325}.

\bibitem[Textile Exchange(2024)]{textileexchange_taxonomy_2024} \href{https://textileexchange.org/knowledge-center/reports/supply-chain-taxonomy-for-the-textile-apparel-and-fashion-industry/}{Textile Exchange (2024)} Supply Chain Taxonomy for the Textile, Apparel, and Fashion Industry.

\bibitem[ThredUp(2025)]{thredup_2025_2025} \href{https://www.thredup.com/resale}{ThredUp (2025)} 2025 Resale Market and Consumer Trend Report.

\bibitem[Tyler et al.(2006)]{tyler_supply_2006} \href{https://doi.org/10.1108/13612020610679295}{Tyler, D. and Heeley, J. and Bhamra, T. (2006)} Supply chain influences on new product development in fashion clothing {\em Journal of Fashion Marketing and Management 10 316-328}.

\bibitem[Watkins and Dunne(2015)]{watkins_functional_2015} \href{https://www.amazon.com/Functional-Clothing-Design-Sportswear-Spacesuits/dp/0857854674}{Watkins, S. and Dunne, L. (2015)} Functional Clothing Design: From Sportswear to Spacesuits.

\bibitem[Wu and Li(2025)]{wu_ai_creativity_2025} \href{https://doi.org/10.1177/00405175241279976}{Wu, J.X. and Li, L. (2025)} AI-driven computational creativity in fashion design: a review {\em Textile Research Journal 95 658-675}.

\bibitem[Wu et al.(2025)]{wu_shiftleft_2025} \href{https://doi.org/10.48550/arXiv.2509.14551}{Wu, X., Li, Z., Hu, F., Lin, T., Zhao, X., Wang, R. and Guo, X. (2025)} Shift-Left Techniques in Electronic Design Automation: A Survey {\em arXiv:2509.14551}.

\bibitem[Yerushalmi and Saha(2025)]{yerushalmi_rebound_2025} \href{https://doi.org/10.1002/bse.70135}{Yerushalmi, E. and Saha, K. (2025)} How Circular Economy Innovation Can Backfire on the Environment: Quantifying the Rebound Effect of the Textiles and Clothing Sector {\em Business Strategy and the Environment n/a}.

\bibitem[Zhang et al.(2024)]{zhang_elasticity_2024} \href{https://doi.org/10.1145/3677388.3696340}{Zhang, J.X., Lin, G.W., Bode, L., Chen, H., Stuyck, T. and Larionov, E. (2024)} Estimating Cloth Elasticity Parameters From Homogenized {Yarn-Level} Models {\em Proceedings of the 17th ACM SIGGRAPH Conference on Motion, Interaction, and Games 1–12}.

\bibitem[Zhou et al.(2025)]{zhou_design2garmentcode_2025} \href{https://doi.org/10.48550/arXiv.2412.08603}{Zhou, F., Liu, R., Liu, C., He, G., Li, Y., Jin, X. and Wang, H. (2025)} Design2GarmentCode: Turning Design Concepts to Tangible Garments Through Program Synthesis {\em arXiv:2412.08603}.

\bibitem[Zhu et al.(2017)]{zhu_unpaired_2017} \href{https://doi.org/10.1109/ICCV.2017.244}{Zhu, J., Park, T., Isola, P. and Efros, A.A. (2017)} Unpaired {Image-to-Image} Translation Using {Cycle-Consistent} Adversarial Networks {\em 2017 IEEE International Conference on Computer Vision (ICCV) 2242-2251}.

\end{thebibliography}
\end{document}